# Using ARIMA to Predict the Expansion of Subscriber Data Consumption

**Mike Wa Nkongolo** [1]

[1] Department of Informatics, Faculty of Engineering, Built Environment and Information Technology, University of Pretoria, Pretoria 0028, South Africa; mike.wankongolo@up.ac.za

**Abstract:** Telecommunication companies face challenges in extracting valuable insights from vast amounts of subscriber data. Predictive analysis can aid in decision-making processes if based on strategic exploration of various data dimensions like geographical, demographic, and financial aspects. This study employs exploratory analysis of 730 data points from Insights Data Storage to forecast subscriber data usage trends using an Auto-Regressive Integrated Moving Average (ARIMA) model. The model yielded a significant p-value of 0.007, supporting the prediction of increased data growth. ARIMA forecasted a 3 Mbps growth with a maximum of 14 Gbps. Compared to the Convolutional Neural Network (CNN), ARIMA demonstrated superior performance, achieving faster execution speeds by a factor of 43. These findings offer insights into predicting subscriber data usage, enhancing Quality of Experience (QoE), and identifying network issues for improved predictive modeling.

**Keywords:** Time series forecasting; Subscriber data; Seasonality; ARIMA; Telecommunication; UGRansome; Stationarity

## 1. Introduction

The growth of competition in the telecommunications industry due to technological variety has facilitated the invention and expansion of new techniques for processing subscriber data to predict their behavior. Subscriber traffic represents all kinds of electronic data transmitted in the network [1]. This data is usually in the form of network flows passing from one node to another [2]. Furthermore, accurately predicting subscriber data can improve the Quality of Experience (QoE) to foresee and predict various anomalies, especially when the company faces revenue loss due to malicious activities. In addition, having the ability to forecast future data usage can be crucial for bandwidth sharing policy within the telecommunication business. Particularly, forecasting integrates a strong sense of seasonality towards data growth to enable management to better predict potential revenue and anomalies.

The above stems from a time series forecasting problem, and there is various research on different forecasting models [3–5]. Statistical models such as Auto-Regressive Integrated Moving Average (ARIMA) and Machine Learning (ML) models such as Long Short Term Memory (LSTM), gradient descent, and regression are popular techniques implemented within the time series forecasting space.

In particular, the LSTM model has demonstrated great forecasting capabilities due to its ability to recall information and it is thus a strong contender against the traditional statistical ARIMA model. Execution speed is another factor to consider when selecting an appropriate model for subscriber data forecasting. This is because subscriber patterns contain petabytes made of historical time series data. It will thus be efficient to consider a model with fast execution speeds to enable faster decision-making.

However, this research addresses the performance of the Convolutional Neural Network (CNN) and ARIMA models to forecast the growth of subscriber data usage. The research determines which algorithm is the most suitable in this scenario and aims to establish which of the two models performs better using speed and accuracy.

In this article, we describe the advantages of using seasonality to examine changes in subscriber data. In a Time Series Analysis (TSA), seasonality is a characteristic of a time series in which the data experiences predictable and regular changes over a period [5]. Understanding seasonality in TSA can enhance the prediction performance of ML models. It can also assist in clearing the features by identifying the seasonality of time series samples and removing them from the original dataset. As a result, one can have a normalized dataset correlating input and output variables.

The seasonality property can also provide more information about the seasonal component of the time series data that can provide insights to enhance predictors' performance [4]. Moreover, modeling seasonality ameliorates the data preparation and feature engineering steps. In each step, seasonal patterns can be extracted and modeled as input/output class labels with a supervised learning scheme.

In adaptive computation, ARIMA is a class of time series forecasting models. Hence it is a special case of a class of regression models, not a class of classification models [6]. We have selected ARIMA as an adequate time series forecasting model to predict subscriber data usage and analyze the seasonality, trends, and cycles of features. The methodology was to use seasonality as the time series data property in the ARIMA model that implemented a distributed lag algorithm to forecast future subscriber data usage based on lagged parameters. This article implements a predictive ARIMA model using subscribers' data to study seasonality by predicting the growth in subscriber throughput.

### 1.1. Research Question

The main research question is as follows:

- Which forecasting model between ARIMA and CNN is effective in predicting subscriber data usage?

The research objective is to evaluate the two models using accuracy and computational speed.

### 1.2. Research Contribution

We propose the ARIMA model for subscriber data prediction using an unsupervised learning scheme. We have specifically implemented the ARIMA model with unlabelled features to predict the growth in subscriber data usage.

In the model, the predictive layer forecasts the throughput rate fed into another layer that predicts the maximum usage growth.

The remainder of the paper is structured as follows. Section 2 discusses related research works and Section 3 the research methodology. Section 4 presents the ARIMA results and the comparative analysis using the UGRansome dataset. Section 5 presents future research directions and concluding remarks.

## 2. Related Works

The section surveys the predictive techniques of TSA with attention to the proposed methodology.

### 2.1. Background

In Ref. [9] a deep learning model to forecast a product usage of a given consumer based on historical data was developed. The authors adapted a CNN with auxiliary input to time-series data to demonstrate an improvement in the model accuracy which predicted future change. To improve the forecasting skills of aircraft in flight navigation systems, Ref. [10] undertook a study on weather forecasting comparing the predictive ability of LSTM and ARIMA models. The study found that the LSTM performs much better than the ARIMA with Root Mean Square Error (RMSE) values of 0.0007 for the LSTM and 0.948 for the ARIMA.

A solution presented by [11] demonstrates that the LSTM outperforms the ARIMA model. The purpose of the study was to forecast a multi-step electricity load for Poland, Italy, and Great Britain. The RMSE values for each model were summarized, but the LSTM outperformed the ARIMA using the RMSE evaluation metric for predicting wind speed.

In Ref. [12], a study to determine which forecasting time series techniques between ARIMA and LSTM produced the most accurate predictions with a minimalistic empirical error was undertaken. The LSTM outperformed the ARIMA for all stock market predictions with an average RMSE of 64 dollars.

Limestone is an important raw material in today's world. Around 10% of the sedimentary rocks on Earth are made up of limestone [13]. According to [13], over 25% of the world's population relies on limestone for drinking water, and about 50% of all known gas and oil reserves are encased in limestone rocks [13]. It is therefore crucial for various economies to accurately predict future prices of limestone. In Ref. [14] a study comparing the ARIMA and LSTM with regards to predicting future prices of limestone was conducted. The ARIMA performed slightly better than the LSTM with an accuracy of 95.7% compared to the LSTM's 91.8%. However, we argue that the probable reason for the LSTM model's subpar performance was due to manual tuning towards some of the model's hyper-parameters.

For instance, the number of LSTM layers was manually tuned. In addition, the author did not disclose the exact units for their target variable. The authors in [15] used regression to learn the correlation between a time series and continuous variables. The approach was to detect the correct coefficients to forecast various attributes. The regression model predicted annual rainfall using historical temperature values [16] with Random Forest (RF) and Gradient Descent (GD) algorithms. The final results confirm the in-depth understanding of time series data to compute the optimal fitting algorithm.

However, Ref. [17] attempted to predict respiratory rates using a sliding window that consists of three modules. The first module retrieved the signal of respiratory patterns; the second approximated the rates, and the third made various estimations. A Gaussian-based regression process extracted the respiratory features from the datasets. It also attempted to fit different Auto-Regressive (AR) algorithms to the retrieved signals. Unfortunately, the AR model failed to detect seasonality.

In Ref. [18], Dynamic Time Warping (DTW) and K-Nearest Neighbor (KNN) used for time series forecasting exhibited a complexity time of 1-NN using the DTW that relied on the engineering of hand-crafted patterns. In Ref. [19], the CNN used on time-series data outperformed all other tested ML models. The author proposed a feature selection method to automate the learning from input variables. The learned patterns represent time series features with discriminatory layers. However, this technique relies on back-propagation that turns the NN into an adequate feature selector.

According to [20], the juxtaposition of Recurrent Neural Networks (RNN) such as LSTM and CNN yielded enhanced accuracy for classification tasks with a range of 27% to 43% in comparison to other well-known ML models. The classification was also considered by [21] and assessed with J48, LSTM, RF, Support Vector Machine (SVM), and CNN. The LSTM-based CNN outperformed other models with three hidden layers.

In Ref. [22], the authors used regression to allocate company resources. In addition, the authors undertook a substantial review of well-known ML models for time series data forecasting, but [23] used CNN to address multivariate time-series regression problems. The LSTM and Gated Recurrent Unit (GRU) portrayed transferable CNN units compared to other models.

The research in [24] used LSTM with additional convolutional layers. The results provide a boost in predicting performance. Lastly, three CNN and four LSTM were implemented by [25] with an improved CNN execution time. Generally, regression models using CNN and LSTM are the most optimal ML techniques used in the literature for time series data forecasting (Table 1). The limitation of the discussed research relies on dataset misunderstanding, lack of feature engineering, non-seasonal patterns, computational biases, and time complexity.

Classifiers such as SVM and Decision Trees (DT) are also prone to error in terms of time series prediction since they are not a better choice for forecasting (Table 1). The time series data forecasting solutions are also implemented in various fields such as weather, electricity, and price prediction (Table 1).

**Table 1.** Comparative analysis.

| Source | Model | Limitation |
|---|---|---|
| [15] | RF&GD | Data understanding |
| [17] | Auto Regressive | Seasonality |
| [22] | CNN | Seasonality |
| [18] | DTW & KNN | Feature engineering |
| [19] | CNN | Back propagation |
| [20] | RNN & LSTM | Classification |
| [21] | LSTM & CNN | Classification |
| [24] | LSTM | Feature engineering |
| [25] | LSTM & CNN | Execution time |
| [23] | CNN | Biases |

**Table 1.** Cont.

| | | Applicability |
|---|---|---|
| [10] | LSTM & ARIMA | Weather forecasting |
| [11] | LSTM & ARIMA | Electricity load forecasting |
| [12] | LSTM & ARIMA | Stock market prediction |
| [14] | LSTM & ARIMA | Limestone price prediction |
| [3] | ABC & ARIMA | Refineries |
| [9] | CNN | Subscriber usage |

## 2.2. Time Series Data Limitation

Some attempts allow efficient computation of large-scale time series data. For instance, Ref. [26] implemented a Hadoop-based framework for accurate preprocessing of data which is important for feature selection. Unlike [26,27] concentrated on model selection by using MapReduce to compute the cross-validation that improved parallel rolling-window prediction using the training set of heterogeneous time series patterns. The predictive parameters computed the accuracy, but this technique could not tackle challenges associated with forecasting.

In Refs. [28,29] multi-step forecasting was monitored by the ML models using the Spark environment. Specifically, Ref. [28] used H iterations to compute the multi-step prediction, while [29] implemented multivariate regression models using ML libraries. As a result, the H technique was not scalable for forecasting. With this, one can use a sample of patterns instead of the original data to predict. For example, Ref. [30] provides an overview of forecasting big data using time series traffic.

The paper provides a premise for time series data forecasting, but it is still complicated to implement the proposed techniques to deal with subscriber data and forecast the future. Some researchers investigate the underlying intuition of parallel computing models using time series data. Unfortunately, these models resulted in expensive computational time complexity.

For instance, Ref. [31] introduced a distributed approximator before the prediction calculation, requiring several iterations. Based on their frameworks, Refs. [32,33] proposed recursive techniques with Bayesian prediction while [34] refined the estimator computation of quantile regression model through various rounds of classification. Another well-known methodology is the alternation of eigenvectors for convex optimization of time series data. This technique blends the seasonality of time series data with the convergence properties of predictors [35], but the streams complicate the forecasting prediction. We argue that a one-shot averaging computation is a straightforward technique to compute the prediction. This method requires only a single computational round [36].

Various studies used distributed learning that split features in a specific frequency domain where the time series patterns are used in the splitting process [37]. These algorithms model successive refinements with a limitation that requires re-implementing each estimator scheme, but slow in terms of convergence accuracy compared to existing predictors designed for time series data forecasting [38].

For example, Ref. [39] analyzed cyclostationary properties of 0-day exploits with slow precision convergence. Boruta was the feature-based extraction method combined with Principal Components Analysis (PCA) to extract the most cyclostationary patterns from NSL-KDD, UGRansome, and KDD99 datasets.

The RF and SVM were used to classify cyclostationary features. The supervised learning restricted the experiments, but our research implements an unsupervised learning scheme to study stationary prediction. Moreover, we have compared the ARIMA performance applied to the UGRansome [40] and subscriber datasets to assess the forecasting performance of stationary and time series data. The following section presents our methodology, Exploratory Data Analysis (EDA), and UGRansome dataset [41]. Lastly, all mentioned articles in this section are crucial because they provide valuable recommendations regarding ML to forecast subscribers' usage data growth.

## 3. Materials and Methods

We have used subscriber data collected from a network database and analyzed the patterns to predict the growth in subscriber data usage. The Network Subscriber Data Management (NSDM) approach is thus the relevant aspect of this research as it stands at the core network layer and stores valuable data used by various subscribers. The NSDM extracts subscribers' patterns from the Insights Data Storage (IDS) and monitors all real-time traffic of subscriber data [42]. We have used the NSDM module that considers subscriber data in a centralized and secure environment having a scalable repository named IDS (Figure 1).

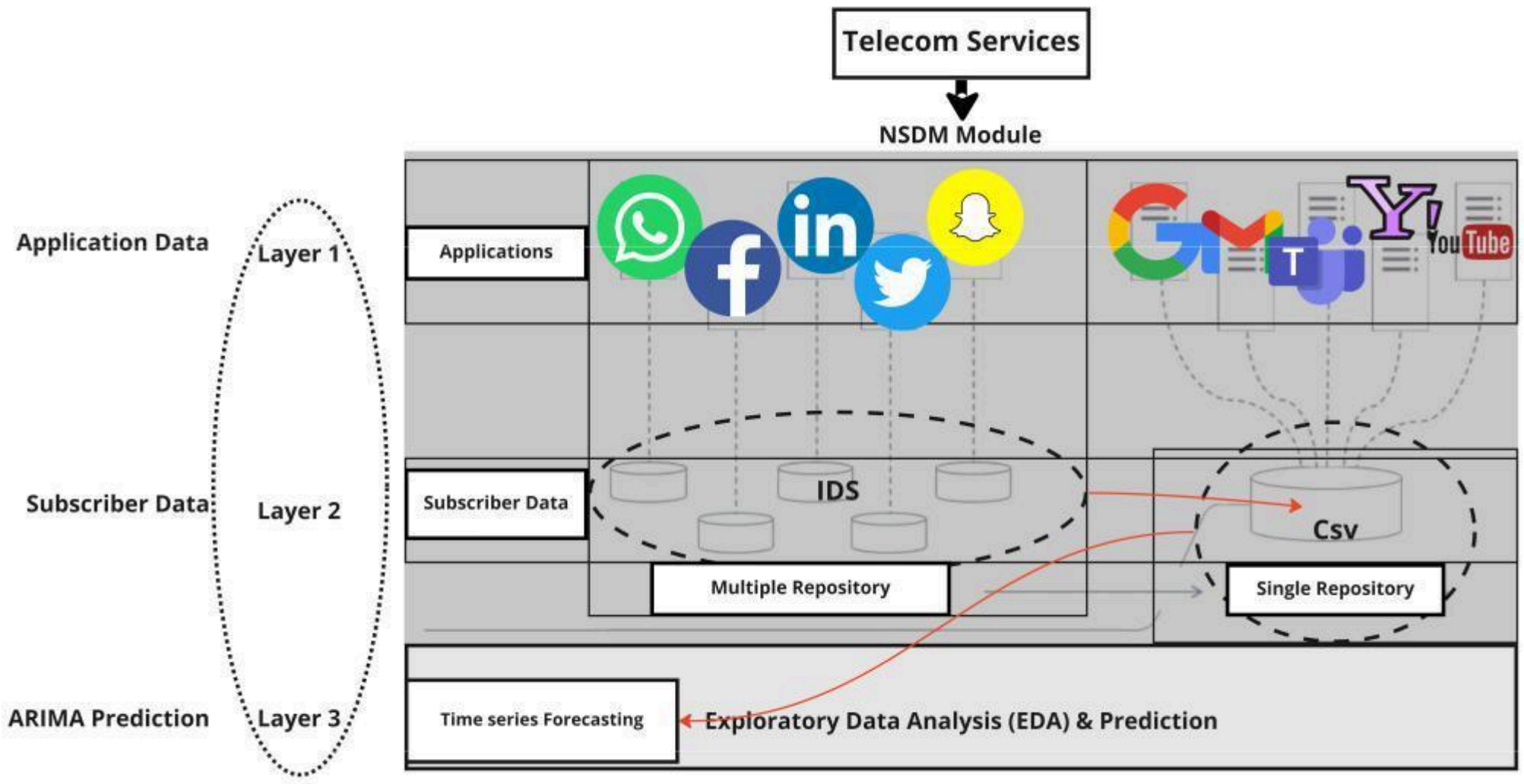

**Figure 1.** The NSDM architecture.

The IDS directory provides distributed and resilient subscriber patterns stored in a single repository. The ARIMA model was used on this repository to predict the growth in subscriber data usage (Figure 1).

### 3.1. ARIMA Formulation

An ARIMA model has a different Moving Average (MA), as well as AR components [43]. We use ARIMA(p, d, q) to denote an ARIMA model where the order of the AR module is (p, q) and d represents the number of differences needed for stationary series [43]. One can extend the ARIMA predictor to a Seasonal ARIMA (SARIMA) model by incorporating additional seasonal patterns to handle time series properties that exhibit a strong seasonal characteristic [43]. We can use ARIMA(p, d, q)(P, D, Q) to formulate a SARIMA model. Here, the uppercase Q, P, and D denote the order of the AR model, the number required for seasonal/stationary series, and the MA order. Similarly, the seasonality period is denoted by m [43,44].

The variation of ARIMA parameters can identify the most optimal set of features in obtaining precise predictive values [43–45].

## 3.2. Experimental Datasets

Figure 2 presents the research methodology where our framework provides subscriber data stored in the IDS module.

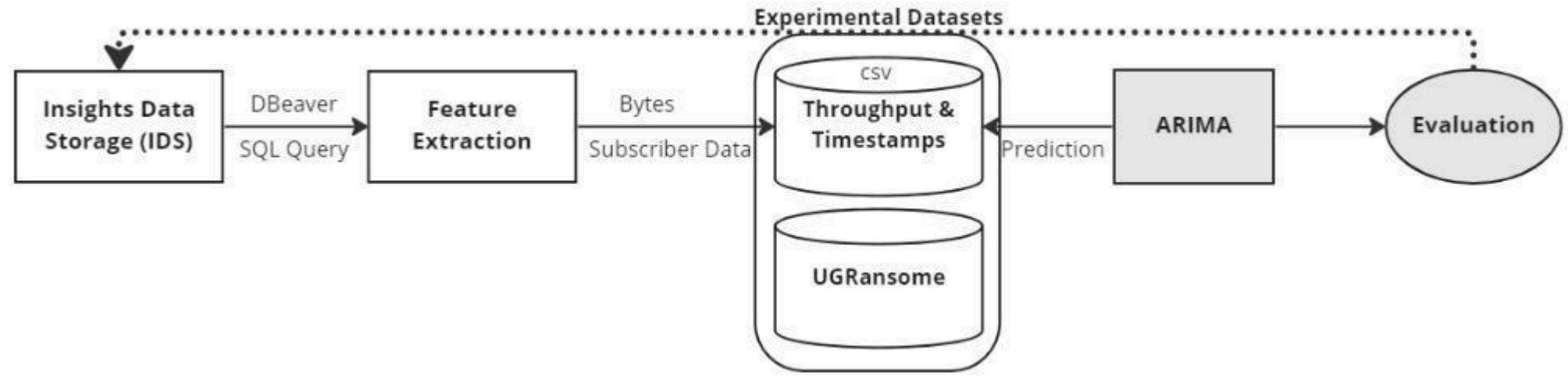

**Figure 2.** The experimental methodology.

The subscriber data was extracted from the real-time network traffic using a Structured Query Language (SQL). We pushed the features into a single comma-separated file and used EDA to visualize salient features of the network traffic. We have then obtained critical Key Performance Indicators (KPIs) that can support the prediction of data usage growth. The executed SQL retrieved the subscriber timestamps, incoming throughput, and outgoing throughout (Figure 3).

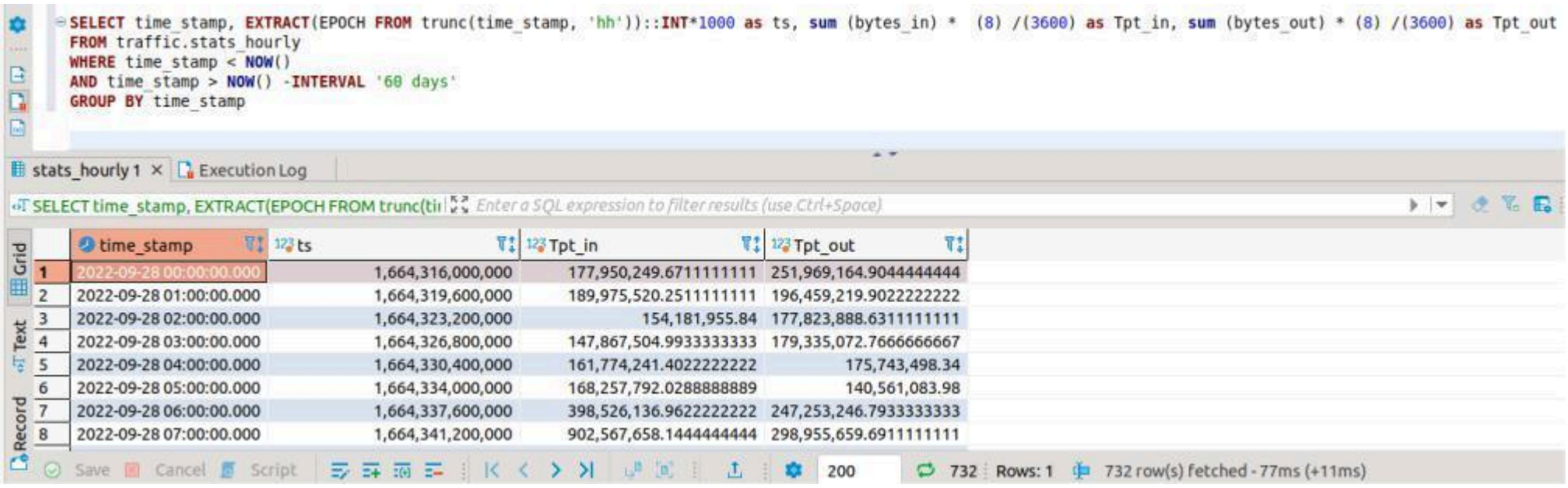

**Figure 3.** The subscriber data.

The query extracts the timestamps (ts) by truncating them into a human-readable format (Year-Month-Time).

The SQL in Figure 3 illustrates this process.

It is hourly-based statistics retrieved from the traffic stats table of the IDS for 60 days (Figure 3). The 3600 represents an hour in seconds, and eight changed the bytes into bits. In addition, we grouped results by timestamps. Retrieved patterns were converted into Comma-Separated Values (CSV) (Figure 4).

The subscriber dataset has 730 entries with four attributes (human-readable times-tamps, UNIX timestamps, incoming throughput (Tpt in), and outgoing throughput (Tpt out)). A timestamp represents the time when the subscriber traffic was collected [46]. The throughput is the flow that measures inputs/outputs movements within the network [46]. The following Figure 5 illustrates our research methodology.

| time_stamp | ts | Tpt_in | Tpt_out |
|---|---|---|---|
| 2022-09-28 0:00:00 | 1664316000000 | 177950249.7 | 251969164.9 |
| 2022-09-28 1:00:00 | 1664319600000 | 189975520.3 | 196459219.9 |
| 2022-09-28 2:00:00 | 1664323200000 | 154181955.8 | 177823888.6 |
| 2022-09-28 3:00:00 | 1664326800000 | 147867505 | 179335072.8 |
| 2022-09-28 4:00:00 | 1664330400000 | 161774241.4 | 175743498.3 |
| 2022-09-28 5:00:00 | 1664334000000 | 168257792 | 140561084 |
| 2022-09-28 6:00:00 | 1664337600000 | 398526137 | 247253246.8 |
| 2022-09-28 7:00:00 | 1664341200000 | 902567658.1 | 298955659.7 |
| 2022-09-28 8:00:00 | 1664344800000 | 2042488462 | 612111154.3 |
| 2022-09-28 9:00:00 | 1664348400000 | 1833562607 | 639776109.2 |

**Figure 4.** The CSV format of the subscriber data.

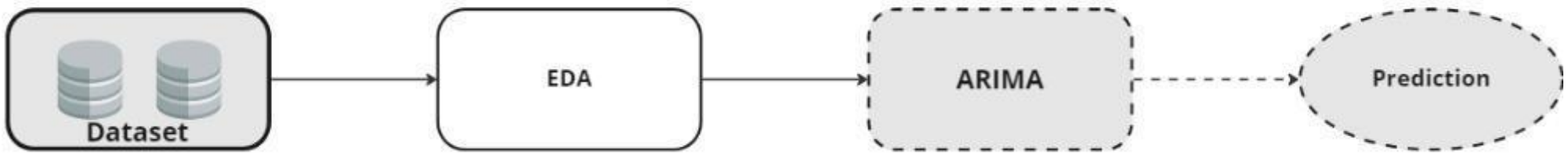

**Figure 5.** The research methodology.

The subscriber and UGRansome datasets are collected, and then the EDA is executed before the computation of the ARIMA model that predicts the growth in data usage based on the current timestamp. The techniques discussed in the literature train ML classifiers with human-labeled features, but this supervised learning method uses limited samples. We have used an unsupervised learning technique whereby we did not label the features. The ARIMA model attempted to use data points $x_1 \ldots x_n$ and assigned predicted values $Q_1 \ldots Q_n$ using predefined parameters.

### 3.3. Feature Engineering and Data Cleaning

Data cleaning is a method of mapping and transforming features from one-row data format to another, to make it more suitable and valuable for various downstream uses, such as time-series forecasting.

One of the most important data cleaning processes is handling missing values [41]. Fortunately, concerning the subscriber data, the dataset contains no missing values. However, the data still needs to be transformed in other various ways for training and testing, and in this case, this will include:

1. ***Data normalization***. The process of normalization is frequently used to prepare data for ML. The objective of normalization is to convert numerical columns to a common scale without losing information or distorting the ranges of values. This will reduce bias towards accurate prediction [40]. We have used Python Sklearn's MinMaxScaler function to normalize the throughput column down to values between 0 and 1.

2. ***Feature engineering***. Also known as feature extraction, is a process of selecting and transforming the most important features from the data to utilize for developing predictive models using statistical or ML models [39]. Concerning the subscriber dataset, only the throughput and timestamp columns will be used to model the training and testing sets.

3. ***Train-test split***. It is a method for validating models that enables one to simulate how a model would behave when presented with fresh untested data [47]. In our analysis, the training data will be split into k-fold cross-validation to avoid under/over-fitting. However, cross-validation is not necessary when the data is small.

### 3.4. Stationarity of Data

Two main methods can be used to determine the stationarity of a time series dataset:

1. ***Visual and graphical inspection***. This is implemented by plotting the functions of the time-series dataset, and then inspecting visually whether the dataset is stationary or not, but the method is prone to inaccuracy.
2. ***Statistical Augmented Dickey-Fuller test***. Named after famous statisticians David Dickey and Wayne Fuller, the Dickey-Fuller test is a more accurate stationarity test method that determines if a time series dataset is stationary by calculating the p-value to test the null hypothesis. The Dickey-Fuller null hypothesis is that the data is not stationary. If the p-value is more than 0.05, then there is strong support for the null hypothesis, and thus the time series dataset will be deemed to be non-stationary. Python's stats model library was used to perform this task by importing the fuller functionality.

In this research, the differencing method was applied to make the dataset stationary (Figure 6). The differencing and original data are distinguished in Figure 6.

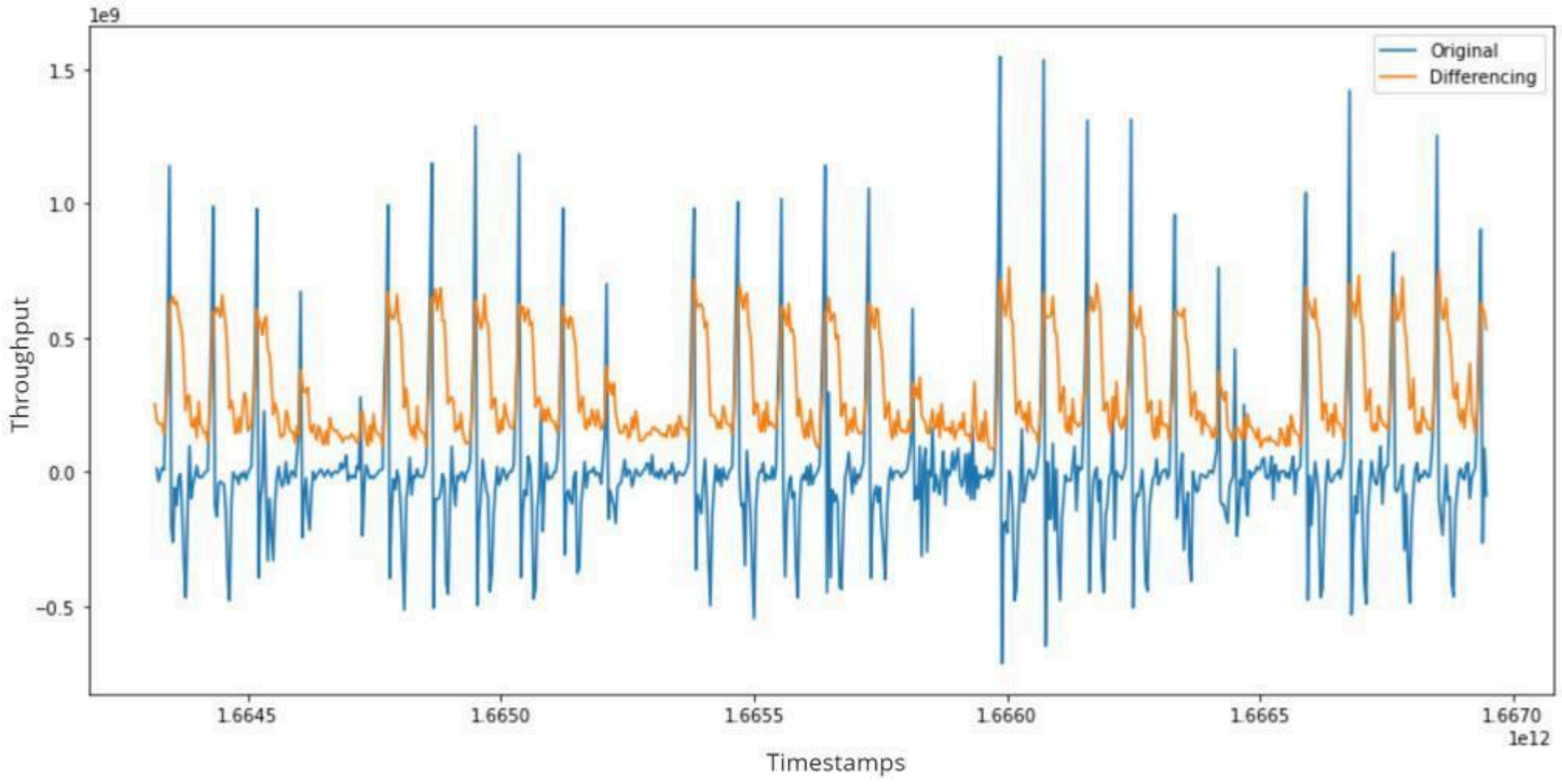

**Figure 6.** The original and differencing of subscriber data.

### 3.5. The UGRansome Characteristics

This dataset was created by extracting important features of two existing datasets (UGR'16 and ransomware) [41]. UGRansome is an anomaly detection dataset that includes normal and abnormal network activities [48]. The regular characteristic sequence makes up 41% of the dataset, whereas irregularity makes up 44%. The remaining 15% represents the predictive values of network attacks grouped into the signature (S), synthetic signature (SS), and anomalous (A) attacks (Figure 7).

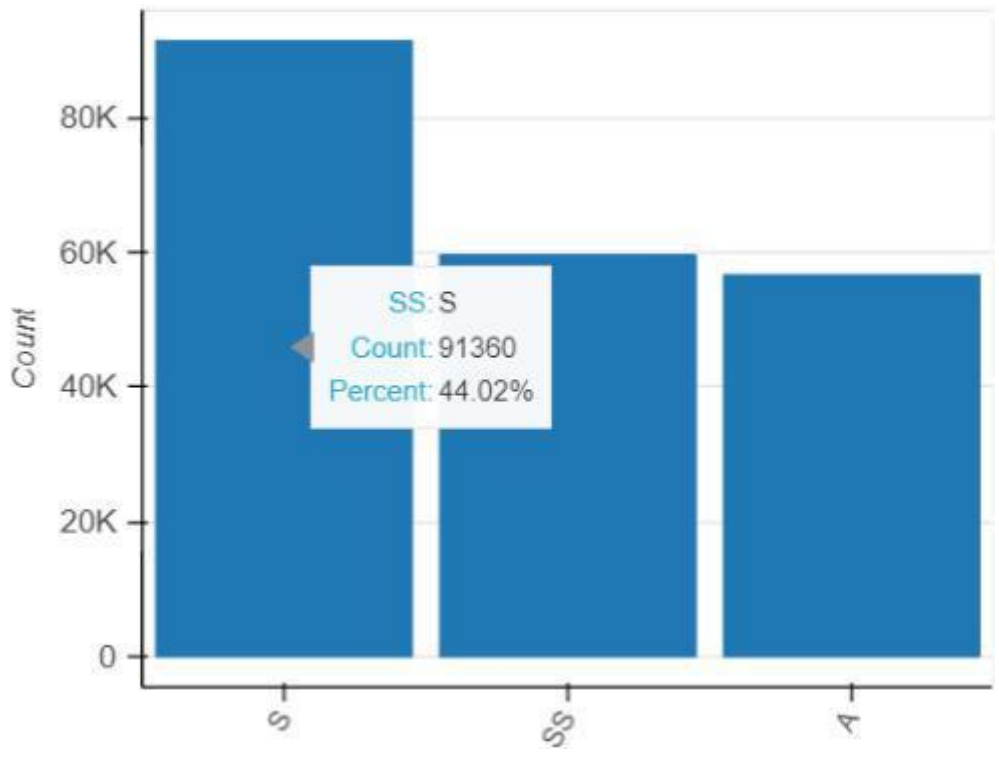

**Figure 7.** Distribution of network threats in the UGRansome dataset.

Figure 7 depicts the signature attacks having a proportion of 44.02% (synthetic signature 28.71%, and anomaly 27.27%). A significant proportion of signature traffic means that the UGRansome threatening concerns are detectable. Regular threats, like User Datagram Protocol (UDP) and Botnet, provide about 9% for the anomalous category. The Internet Protocol (IP) and ransomware addresses have a ratio of 1% [39]. In addition, a ratio of 2% exists between communication protocols and ransomware addresses [41]. According to Refs. [39,41] the significant distribution of the UGRansome could be summed up in the following Figure 8. However, UGRansome is more redundant compared to subscriber data and we removed 28.2% of duplicate records during the feature extraction phase (Figures 8 and 9).

**Dataset Statistics**

| | |
|---|---|
| Number of Variables | 14 |
| Number of Rows | 207533 |
| Missing Cells | 0 |
| Missing Cells (%) | 0.0% |
| Duplicate Rows | 58491 |
| Duplicate Rows (%) | 28.2% |
| Total Size in Memory | 106.9 MB |
| Average Row Size in Memory | 540.2 B |
| Variable Types | Numerical: 4<br>Categorical: 9<br>GeoGraphy: 1 |

**Figure 8.** The UGRansome data summary.

**Dataset Statistics**

| | |
|---|---|
| Number of Variables | 4 |
| Number of Rows | 730 |
| Missing Cells | 0 |
| Missing Cells (%) | 0.0% |
| Duplicate Rows | 0 |
| Duplicate Rows (%) | 0.0% |
| Total Size in Memory | 74.3 KB |
| Average Row Size in Memory | 104.2 B |
| Variable Types | Categorical: 1<br>Numerical: 3 |

**Figure 9.** The subscriber data summary.

### 3.6. Exploratory Techniques

The exploratory analysis provides a set of techniques to understand the dataset. The results produced by the EDA can assist in mastering the data structure [49], as well as the distribution of the features, detection of outliers, and correlation within the dataset. Some of the statistical metrics used to evaluate the ARIMA model are standard deviation, correlation, mean, standardized residual, normal Q-Q, correlogram, theoretical quantile, p-value, and accuracy:

- ***Standardized residual ($r_i$)***. It measures the strength of actual and predicted values and indicates the significance of features [50] ($r_i$ facilitates the recognition of patterns that contribute the most to the predictive values).

- ***Normal Q-Q***. The normal Q-Q means normal Quantile-Quantile. It is a plot that compares actual and theoretical quantiles [50]. The metric considers the range of random variables to plot normal Q-Q using a probabilistic computation. The x-axis represents the Z-score of the standardized normal distribution, but different formulations have been proposed in the literature to detect the plotting positions.

- ***Correlogram***. It is a correlational and statistical chart used in TSA to plot the auto-correlations sample $r_h$ versus the timestamp lags h to check for randomness [50]. The correlation is zero when randomness is detected.

- ***Augmented Dickey-Fuller (ADF) test***. This statistical metric tests the stationarity of time series data [50] by using a unit root metric *b* that exists in a series of observations.

- ***Theoretical quantile***. The theoretical Q-Q explores the variable's deviation from theoretical distributions to visually evaluate if the ratio is significantly different for EDA purposes [50].

- ***Kurtosis***. This metric evaluates the probability of the predicted variables by describing the probability proportion. There are various techniques to compute the theoretical distribution of Kurtosis, and there are subjective manners of approximating it with relevant samples [50]. With Kurtosis results, higher values determine the presence of outliers.

- ***Jarque-Bera (JB) test***. This metric uses a Lagrange multiplier to test for data normality. The JB value tests if the distribution is normal by testing the Kurtosis to determine if features have a normal distribution. A normal JB distribution will have symmetrical

Kurtosis indicating the peak in the distribution.

- ***Heteroscedasticity***. It checks the alternative hypothesis ($H_A$) versus the null hypothesis ($H_0$) [50]. With the alternative hypothesis, the empirical error is multiplying the function of various variables:

$$H_A : s_1^2 = s_2^2 \; ... \; s_n^2.$$

However, a null hypothesis has equal error variances (homoscedasticity) [50]:

$$H_0 : s_1^2 = s_2^2 = ... = s_n^2.$$

- ***Accuracy***. The balanced accuracy $B_A$ of the ARIMA model is calculated with the following mathematical formulation [47]:

$$B_A = \frac{((TP/TP + FN) + (TN/(TN + FP)))}{2}$$

where True Positive (TP) and True Negative (TN) denote correct classification, but misclassification is the False Positive (FP) and False Negative (FN) [50]. We used cross-validation rounds to build multiple training/testing subsets to decide which model is a suitable predictor of the growth in subscriber data (80% of the training set, 10% of the validation set, and 10% of the testing set).

### 3.7. Feature Extraction

ML models are used to address a range of prediction problems. The unsatisfactory prediction of ML classifiers originates from overfitting or underfitting features [7]. In this research, the removal of irrelevant patterns guaranteed improved performance of the ARIMA computation. PCA was utilized on the UGRansome data to extract relevant patterns. PCA is a feature extraction methodology of this research. It uses stochastic x(t) and t = 1, 2 . . . l with n-dimensional inputs x having a probability matrix **P** of zero mean. The PCA formulation uses the covariance with a linear calculation of x(t) inputs into y(t) outputs:

$$y(t) = Q^T x(t),$$

Q is an orthogonal n*n matrix type where i represents the columns viewed as eigen-vectors computed as follows:

$$y_i(t) = Q^T x(t),$$

The range in Equation (20) starts from 1 . . . n where $y_i$ is the new component of the ith PCA. Table 2 depicts the PCA results using the UGRansome dataset. However, SQL was used as a feature extractor to extract subscriber data from the IDS (Figure 3).

**Table 2.** The PCA results using the UGRansome.

| Attack | Feature | Total |
|---|---|---|
| Blacklist | Timestamp | 2761 |
| Spam | IP address | 7425 |
| Scan | Flag | 1559 |
| SSH | Prediction | 7293 |
| Botnet | Threats | 4765 |
| Total | - | 23,803 |

### 3.8. Model Training and Testing

The parameters shown in Figure 10 will be used to train and test the ARIMA model.

| | p | d | q |
|---|---|---|---|
| Subscriber data | **2** | **1** | **2** |
| UGRansome data | **1** | **2** | **1** |

**Figure 10.** The ARIMA model parameters.

The choice of these parameters is justified because a model with only two AR terms would be specified as an ARIMA of order (2, 0, 0). A MA(2) model would be specified as an ARIMA of order (0, 0, 2). A model with one AR term, a first difference, and one MA term would have order (1, 1, 1). For the proposed model using the subscriber data, an ARIMA (2, 1, 2) model with two AR terms, the first difference, and two MA terms are being applied to improve the forecasting accuracy (Figure 10). One AR(1) term with two differences and one MA(1) term are used for the UGRansome data to account for a linear trend in the data (Figure 10).

The differencing order refers to successive first differences. For instance, for a difference order = 2 the variable analyzed is $z_t = [x_t \; x_{t+1}] \; [x_{t+1} \; x_{t+2}]$. This type of difference might account for a quadratic trend in the data. However, due to seasonality concerning the experimental data, the SARIMA model is used. The SARIMA model is an extension of the ARIMA with integrated seasonal components. The training set is used to train the model and thus used to predict the data usage for the 2022 year. The test set is also utilized to validate the final predictions for the year.

### 3.9. Model Tuning

In contrast to the ARIMA, the CNN has several parameters that are not estimated by the model (i.e., the p and q values), and thus the algorithm is trained by manually specifying a set of hyper-parameters using trial and error. The number of layers, neurons, batch sizes, and epochs utilized in the CNN, are some examples of hyper-parameters that need to be tuned manually. It should also be noted, that, unlike the ARIMA which required the dataset to be transformed to its stationary format, the CNN in this research will be trained against non-stationary time series patterns of the UGRansome dataset. Considering a single input of features as depicted in Figure 11, the CNN weights these features to enable the learning trend of particular observations. We have various observations, so we merge their outputs into a hidden layer.

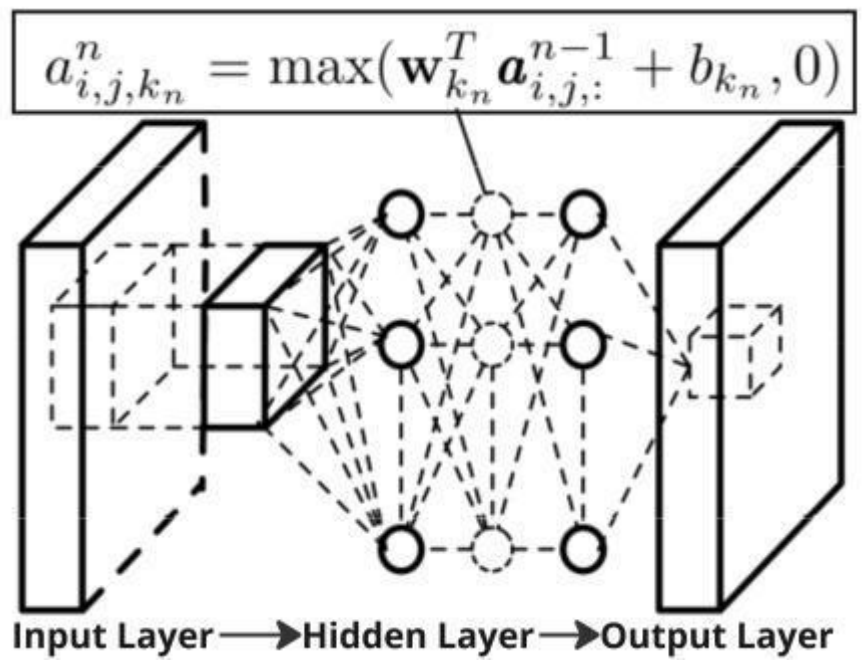

**Figure 11.** The CNN architecture.

Our CNN architecture uses binary convolutions (with 70 and 30 filters) and a densely connected layer of 50 neurons with the activation function (ReLU). This unit has six connected layers representing auxiliary outputs (Figure 11). Each layer predicts and passes the prediction value to the next layer which predicts growth in the subscriber data usage until the final layer produces long-term forecasting.

We use each layer to predict in advance some additional days and grid search to detect the optimal number of filters, convolutions, connected layers, and drop-out rate. For each layer k [1, 2, 3, 4, 5, 6] we added an empirical error and loss function. Each layer k aims to produce future forecasting for more than 14 days ahead:

where [$y_i$, $14_k$] represents the features of time-series i, and [$\hat{y}_i$, k] the k-th layer's forecasting the value. The function restricting the range of [$\hat{y}_i$, k] to [ 1, 1] is denoted by *tanh*.

Table 3 shows a summary of the hyper-parameters used for the CNN model.

**Table 3.** CNN tuning parameters.

| Hyper-Parameter | Value |
|---|---|
| Number of neurons | 50 |
| Activation function | Tanh |
| Number of dense layers | 6 |
| Optimizer | Adam |
| Batch size | 2 |
| Loss function | Mean Squared Error (MSE) |
| Number of epochs | 50 |

### 3.10. ARIMA Predictor Model

Given a long period, $y_t$[t = 1, 2, . . . , T] of a spanning time series traffic, the aim is to come up with a new scheme that works well for predicting the future outcomes H. We define S = [1, 2, . . . , T] as the sequence of timestamp with time series $y_t$. The estimation was merged before the prediction. The idea is to use g(.) as a single mean parameter, and our computational framework could be viewed as an averaging ML algorithm (Figure 12).

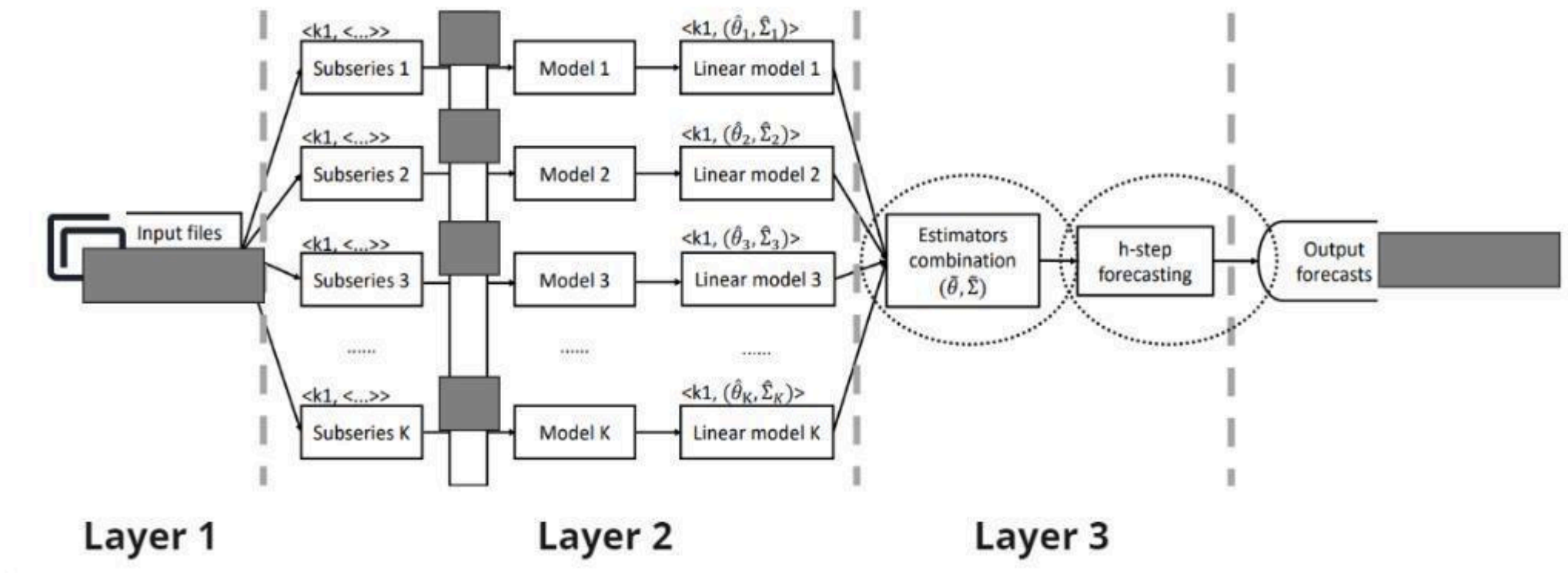

**Figure 12.** The ARIMA model.

Figure 12 outlines the proposed ARIMA model to forecast the growth in subscriber data. The timestamps of historical data were recorded in the IDS before being processed by the ARIMA. In simple terms, the proposed model consists of the following phases:

- ***Phase 1***: Preprocessing. Subdivide the time series data into the K sub-series.
- ***Phase 2***: Modeling. Train the algorithm using sub-series data by assuming that the IDS of the sub-series remains constant.
- ***Phase 3***: Linear transformation. Translate the trained algorithm in phase 2 into linear representations k.
- ***Phase 4***: Estimator combination. The obtained local estimator from phase 3 minimizes global losses parameters described in Section 3.1.
- *Phase 5*: Prediction. Predict the next observations H by utilizing the merged estimator's parameters presented in Equations (24) and (25).

We used available hourly-based timestamps to create a new set of timestamps (ts) used in the ARIMA prediction. The following formulation was used to predict new timestamps (Equation (26)):

$$\text{Predicted}_{(ts)} = \text{Last_EpochTimestamps} + n * 3600$$

where n = range(1, 48). This computation provides new predicted timestamps on an hourly basis for the next hours. Figure 13 shows the computation of the needed inputs x using the current value, step size, and stop value. The x represents current hourly values starting at zero milliseconds (ms) and going up to 731. The computation of the predicted values used x, a step size of 360,000, and a stop value. The needed value represents the result of the prediction in which the current value of x was used as an index starting at zero and multiplied by the step size. This calculation stops at the 731st iteration, where

731 denotes the last index
(Figure 13).

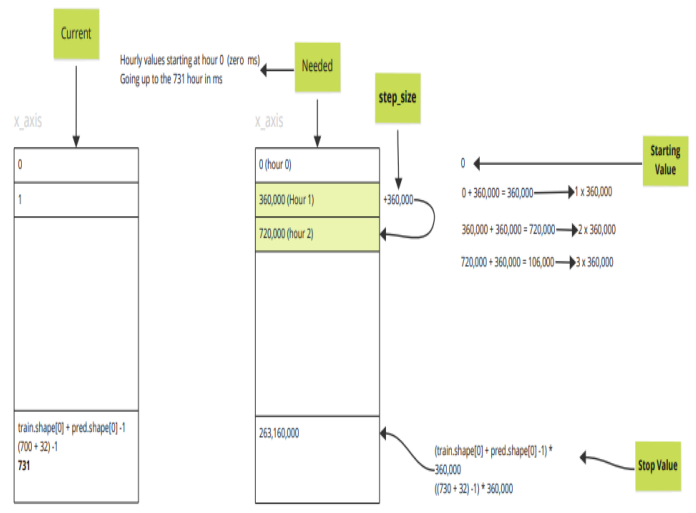

**Figure 13.** The predictive computation.

3.11. Computational Environment

The IDS used to build the subscriber data is installed on a DBeaver database. DBeaver is a database monitoring software that manipulates subscriber data. It can be used to build analytical dashboards from various data storage. Table 4 presents the computing environment and Table 5 the feature extraction results.

**Table 4.** Framework specification.

| Node | Specification |
|---|---|
| RAM | 39 GB |
| Service | Jupyter & DBeaver |
| ML algorithm | ARIMA & CNN |
| System | 64-bits |
| Processor | 2.60 GHz |
| Dataset | Subscriber data & UGRansome |
| Operating System (OS) | Windows & Linux |
| CPU | Intel i7-10 |
| Language | Python & SQL |

3.12. Feature Extraction

There are different reasons causing duplication in a dataset, among which are imperfections in the data collection process and the properties of patterns, but feature extraction solves redundancy dimensions. Features projected into a new space have lower dimensionality.

Examples of such techniques include Linear Discriminant Analysis (LDA), Canonical Correlation Analysis (CCA), and PCA [40]. We have used PCA to extract relevant features of the UGRansome dataset. The PCA lowered computational complexity, built generalizable models, and optimized the storage space. To address the redundancy issue, the PCA selected a subset of relevant patterns from the original dataset based on their relevance.

We present the PCA results in Table 5 where the final dataset with the description of each attribute is presented. The prediction attribute facilitates the forecasting of any ML model to predict the category of novel intrusion. Our final dataset has 12 variables with 180,564 observations (Table 5).

Consequently, if the deviation degree of a variable is high or low enough, it is considered an abnormality. However, we did not apply feature extraction on the subscriber data because it has no redundant patterns. The feature extraction using UGRansome leads to improved performance, higher prediction accuracy, minimized computational time, and efficient model interpretability.

**Table 5.** Extracted UGRansome features.

| Number | Attribute | Description | Type |
|---|---|---|---|
| 1 | Timestamp | Traffic duration | Numeric |
| 2 | Protocol | Communication rule | Categorical |
| 3 | Flag | Network state | Categorical |
| 4 | IP address | Unique address | Categorical |
| 5 | Network traffic | Periodic network flow | Numeric |
| 6 | Threat | Novel malware | Categorical |
| 7 | Port | Communication port | Numeric |
| 8 | Expended address | Malware address | Categorical |
| 9 | Seed address | Malware address | Categorical |
| 10 | Cluster | Group assigned | Numeric |
| 11 | Ransomware | Novel malware | Categorical |
| 12 | Prediction | Novel malware class | Categorical |

## 4. Results

This section is structured into the DFT, comparative results of ARIMA and CNN, execution speed test, and predictive results comparison using additional standard forecasting approaches such as BATS and TBATS. The section compares the ARIMA model performance using the subscriber and the UGRansome datasets. Figure 14 shows the format of the subscriber data following a stationary distribution compared to the UGRansome timestamps.

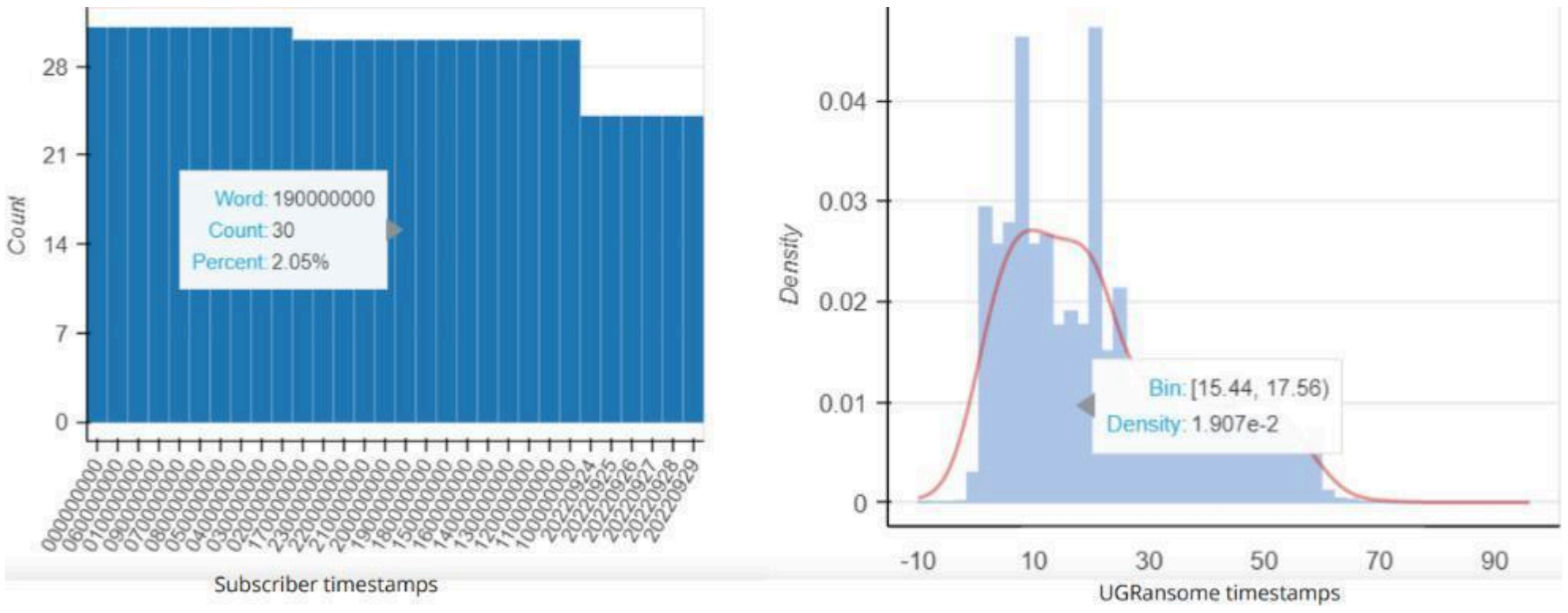

**Figure 14.** The timestamp and density comparison.

Figure 15 portrays a distribution of incoming and outgoing throughput of the subscriber data compared to the UGRansome port traffic (5066–5068).

Each attack flow is also depicted. The figure depicts NerisBotnet threats with less traffic. This result reveals a time series forecasting property of the subscriber data but not the UGRansome. However, the UGRansome has more distributed or dependent variables (Figure 16).

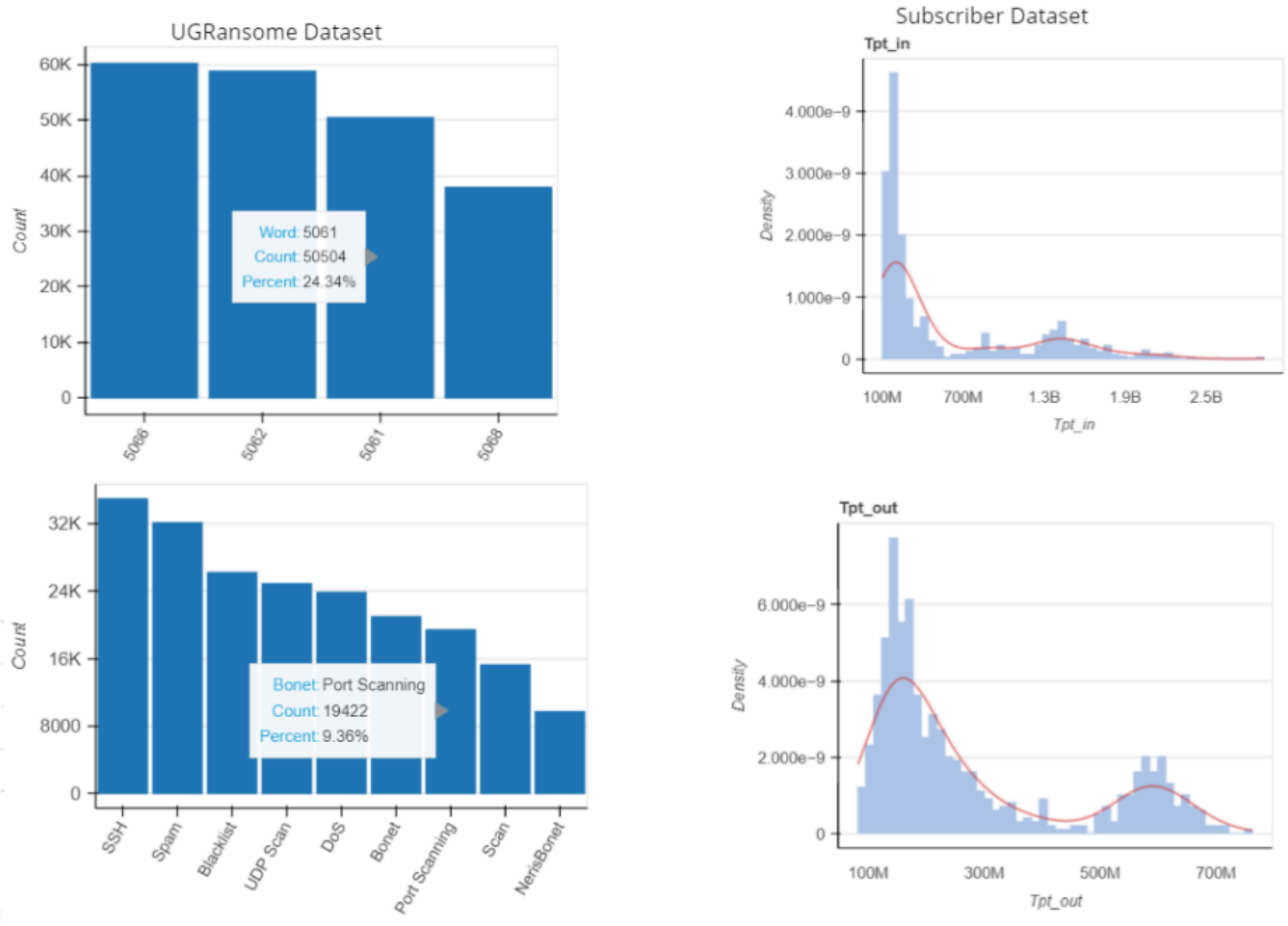

**Figure 15.** Additional features comparison.

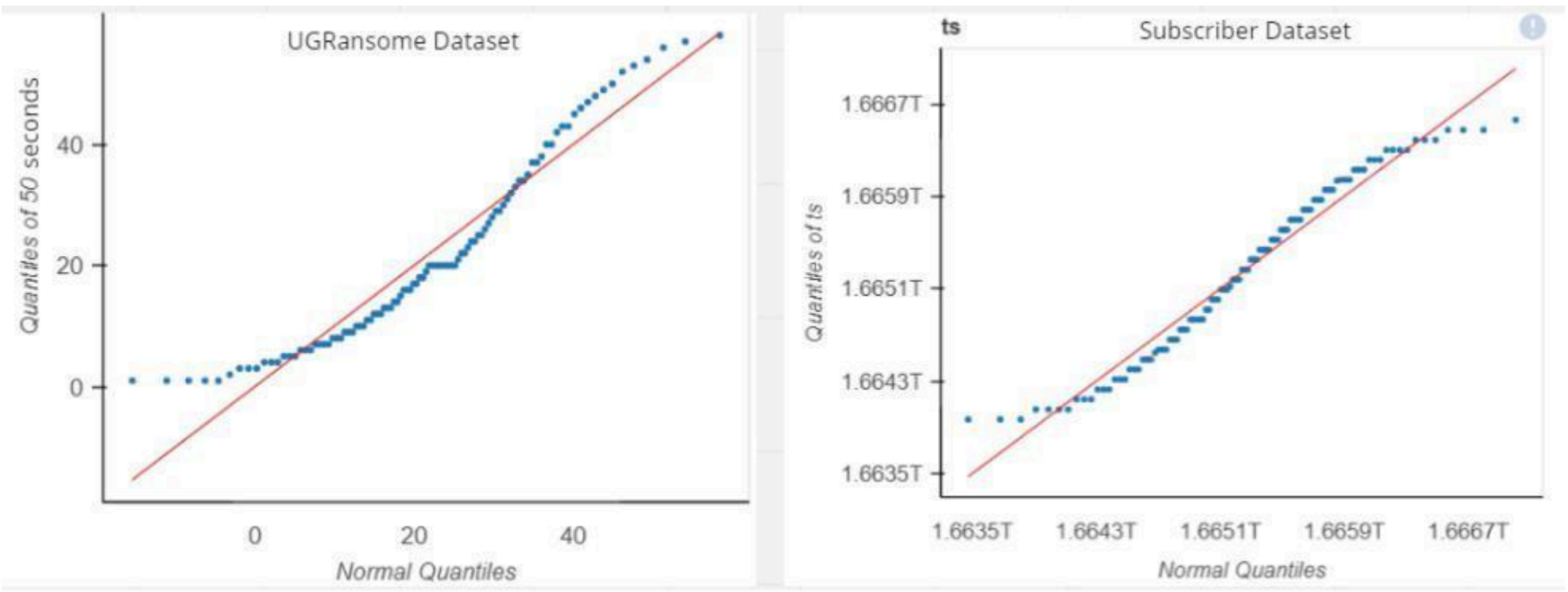

**Figure 16.** The normal Q-Q results.

The correlation of throughput is in Figure 17.

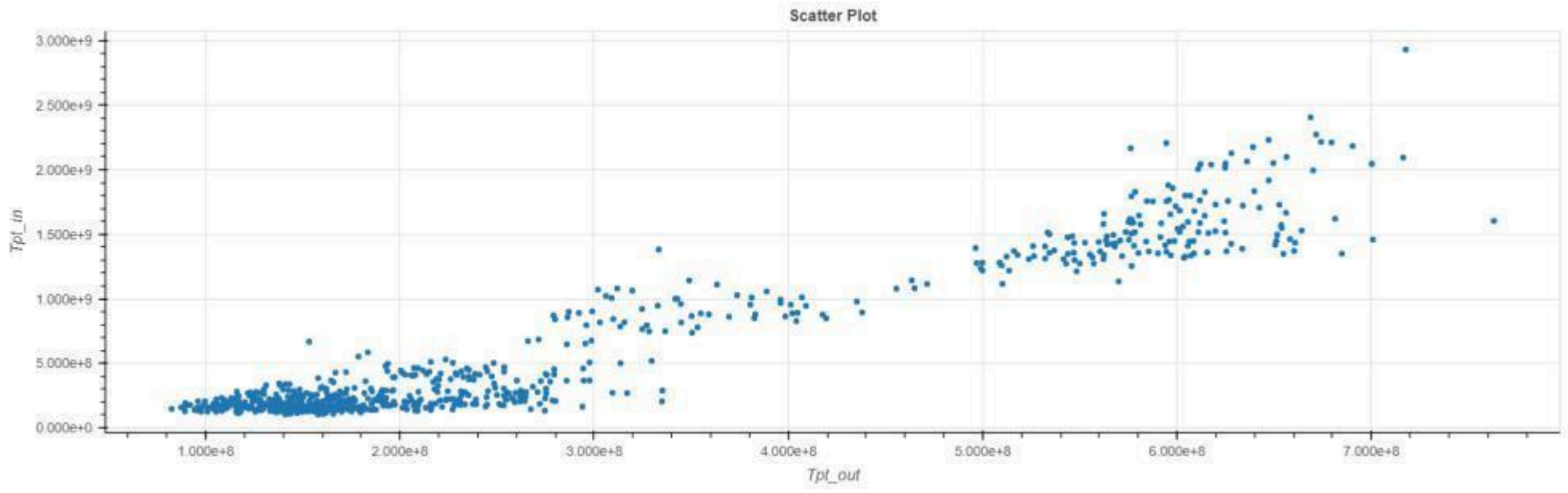

**Figure 17.** The throughput correlation.

This plot indicates the linear distribution of predicted values. The summary of the SARIMA model using the subscriber data is presented in Figure 18.

SARIMAX Results

| | | | |
|---|---|---|---|
| Dep. Variable: | y | No. Observations: | 732 |
| Model: | SARIMAX(3, 0, 1) | Log Likelihood | -14399.470 |
| Date: | Mon, 31 Oct 2022 | AIC | 28810.941 |
| Time: | 17:51:26 | BIC | 28838.515 |
| Sample: | 0 | HQIC | 28821.578 |
| | - 732 | | |
| Covariance Type: | opg | | |

| | coef | std err | z | P>\|z\| | [0.025 | 0.975] |
|---|---|---|---|---|---|---|
| intercept | 4.899e+07 | 5.38e-09 | 9.1e+15 | 0.000 | 4.9e+07 | 4.9e+07 |
| ar.L1 | 1.1487 | 0.103 | 11.154 | 0.000 | 0.947 | 1.351 |
| ar.L2 | -0.1228 | 0.135 | -0.912 | 0.362 | -0.387 | 0.141 |
| ar.L3 | -0.1855 | 0.051 | -3.621 | 0.000 | -0.286 | -0.085 |
| ma.L1 | -0.1868 | 0.118 | -1.588 | 0.112 | -0.417 | 0.044 |
| sigma2 | 7.181e+15 | 3.07e-17 | 2.34e+32 | 0.000 | 7.18e+15 | 7.18e+15 |

| | | | |
|---|---|---|---|
| Ljung-Box (L1) (Q): | 0.02 | Jarque-Bera (JB): | 376.22 |
| Prob(Q): | 0.88 | Prob(JB): | 0.00 |
| Heteroskedasticity (H): | 1.35 | Skew: | 1.22 |
| Prob(H) (two-sided): | 0.02 | Kurtosis: | 5.54 |

**Figure 18.** The SARIMA model summary.

The summary confirms that the prediction of subscriber data will have an increased mean or standard deviation given the likelihood, Prob(Q), and Kurtosis values. The SARIMA model reports no outliers for subscriber data given the Kurtosis value, and the JB test describes a normal distribution. Similarly, the level of heteroscedasticity supports our hypothesis with a degree of 1.35, a p-value of 0.112 having a likelihood value of 14,399.47. With these results, we reject the null hypothesis and accept the alternative. The former denies the growth of subscriber data, while the latter accepts. In the next section, we will compare the subscriber data with the UGRansome using the DFT results.

## 4.1. Dickey Fuller Test

The DFT results are in Table 6. A p-value of 0.007 with an accuracy of 90% was obtained. This result supports our hypothesis predicting an increase in subscriber data growth. As such, (i) the UGRansome used more iterations due to its size surpassing the subscriber dataset, (ii) the balanced accuracy of the DFT reached 81% of accuracy, and (iii) the dataset size had no effect on the prediction performance. The residual and correlogram are in Figure 19 with the seasonality of the throughput.

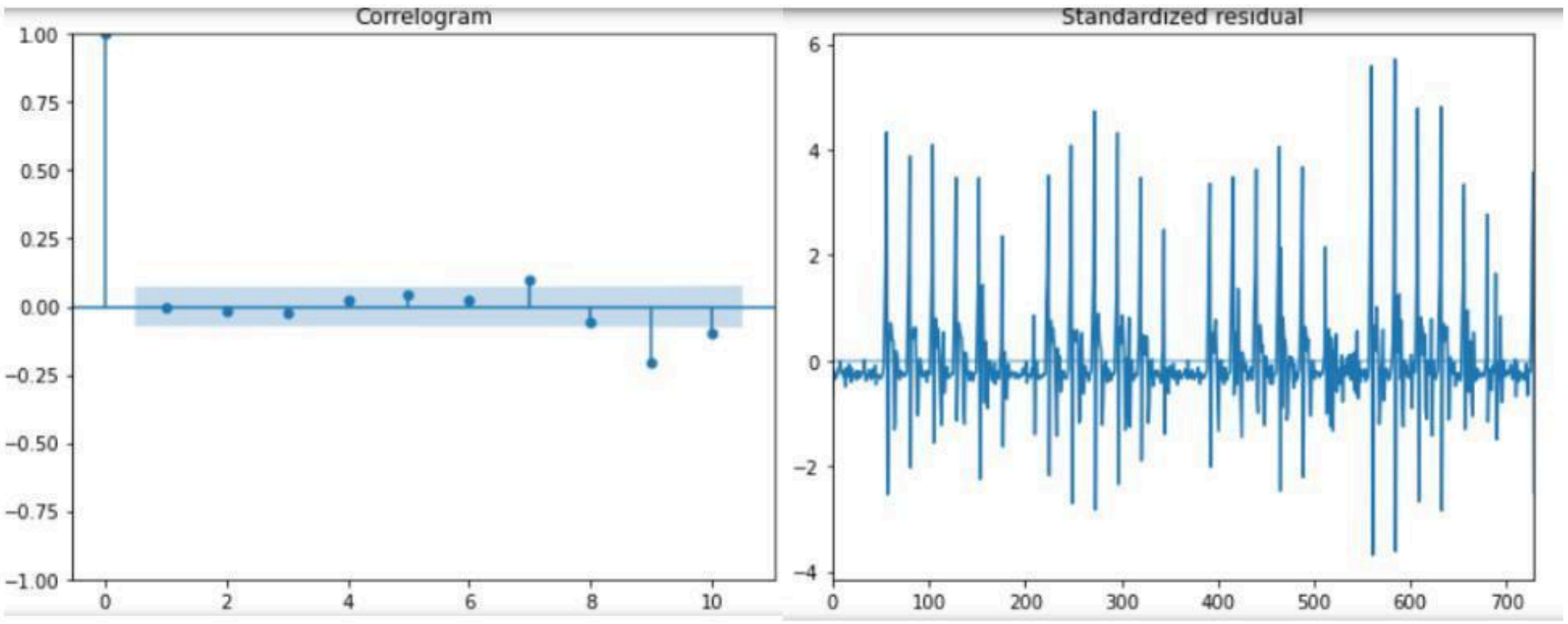

**Figure 19.** The standardized residual and correlogram.

**Table 6.** The DFT results.

| Dataset | Test Statistic | p-Value | Iteration | Accuracy |
|---|---|---|---|---|
| Subscriber data<br>UGRansome data | 3.537879<br>9.876982 | 0.007066<br>0.0008044 | 20<br>342 | 90.567%<br>90.456% |
| | **Correlogram** | **ADF** | **Q-Q** | |
| Subscriber training set | 0.9 | 0.8 | 0.9 | 90.398% |
| UGRansome training set | 0.8 | 0.9 | 0.7 | 89.453% |
| Subscriber testing set | 0.8 | 0.9 | 0.9 | 91.348% |
| UGRansome testing set | 0.8 | 0.8 | 0.9 | 88.298% |
| | **Features Total** | **Mean** | **Deviation** | |
| Subscriber data | 700 | 54.23 | 22.45 | 92.351% |
| UGRansome data | 8932 | 75.32 | 46.3 | 88.527% |
| Subscriber testing set | 400 | 12.6 | 6.7 | 94% |
| UGRansome testing set | 4765 | 26.87 | 39.65 | 88% |
| **Balanced Accuracy** | - | - | - | **81%** |
| **Balanced Features** | **3699** | - | - | - |
| **Balanced Mean** | - | **41.75** | - | - |
| **Balanced Deviation** | - | - | **28.25** | - |

However, the data usage growth prediction is illustrated in Figure 20 using the ARIMA model that predicts a growth of 3 Mbps at a specific timestamp. UNIX timestamps of the subscriber data predicted the maximum data growth using the ARIMA model.

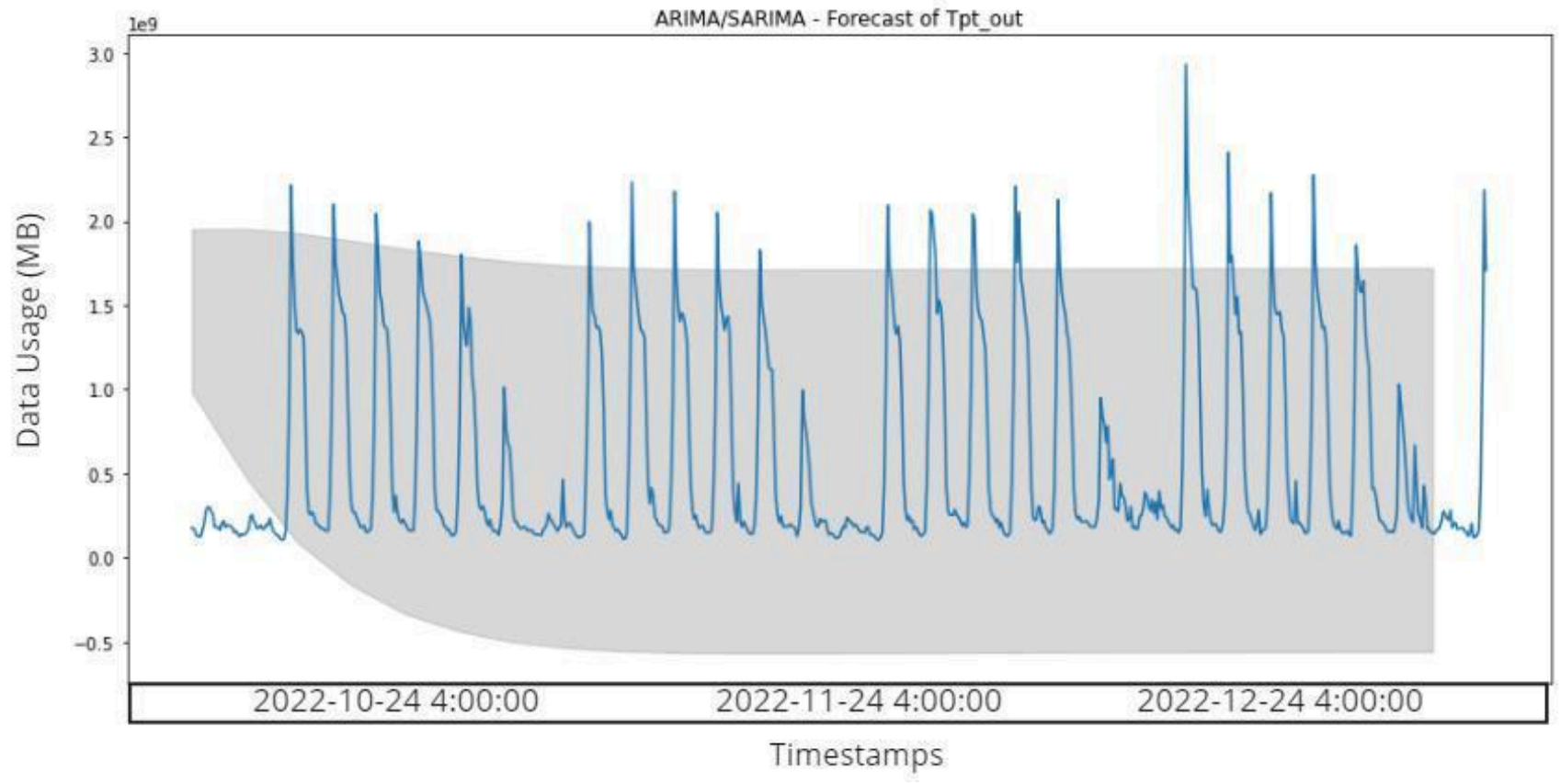

**Figure 20.** The ARIMA prediction.

In Figure 21, ARIMA predicts a maximal subscriber data usage growth of 14 Gbps (where blue denotes actual data, orange is the predicted ARIMA data, and green is the future predicted values). The predicted mean and standard deviation are in Figure 22.

The original data represents current values, but ARIMA approximated a mean and standard deviation from these values. ARIMA predicted a maximum mean value of 5 Mbps with a standard deviation of 2 Mbps. The results show mean and standard deviation values lower than the original or current values. The scatter plot of the testing set is shown in orange with a testing size of 32 data points that have been plotted or predicted (Figure 23).

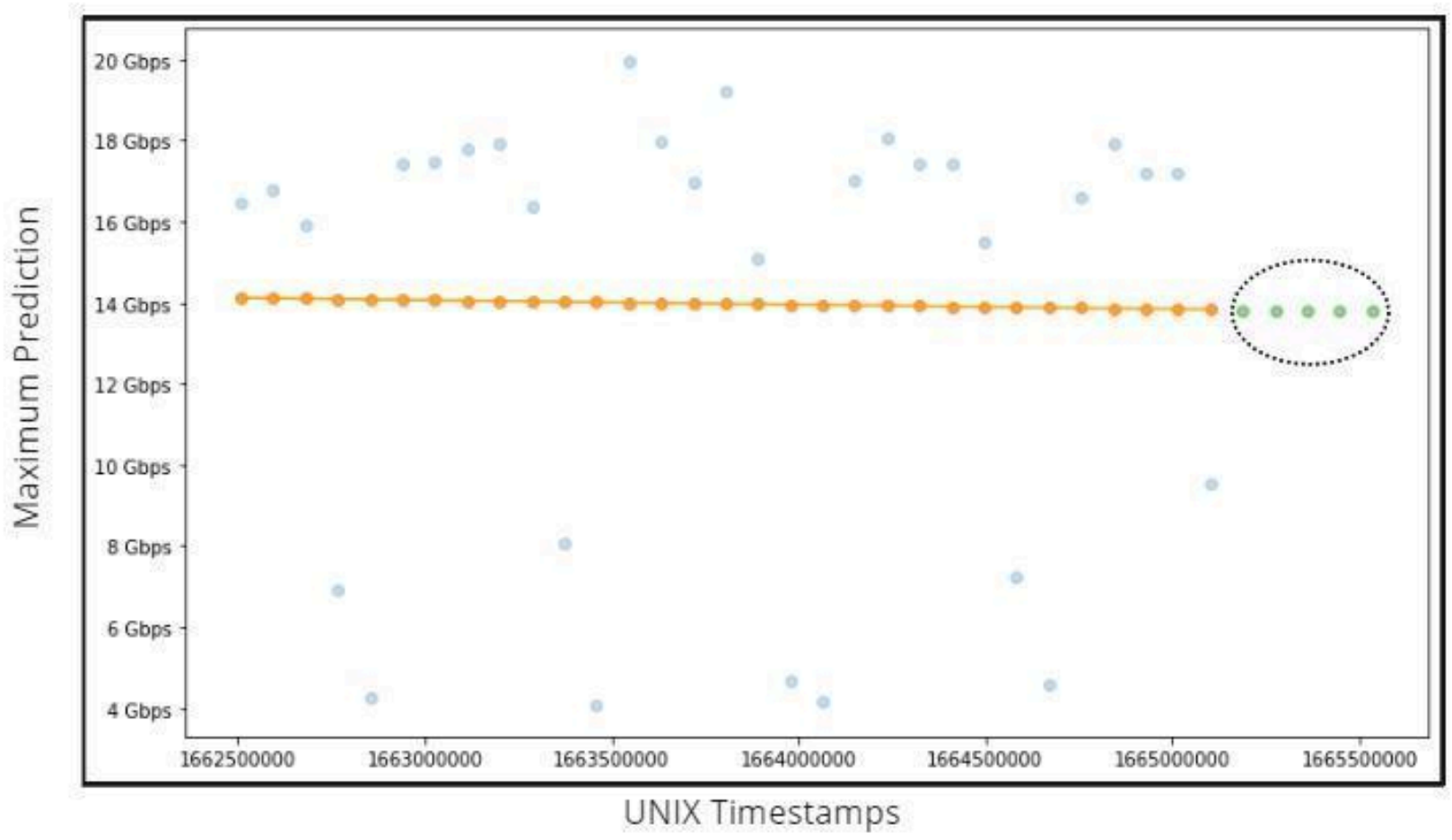

**Figure 21.** The maximum data usage prediction.

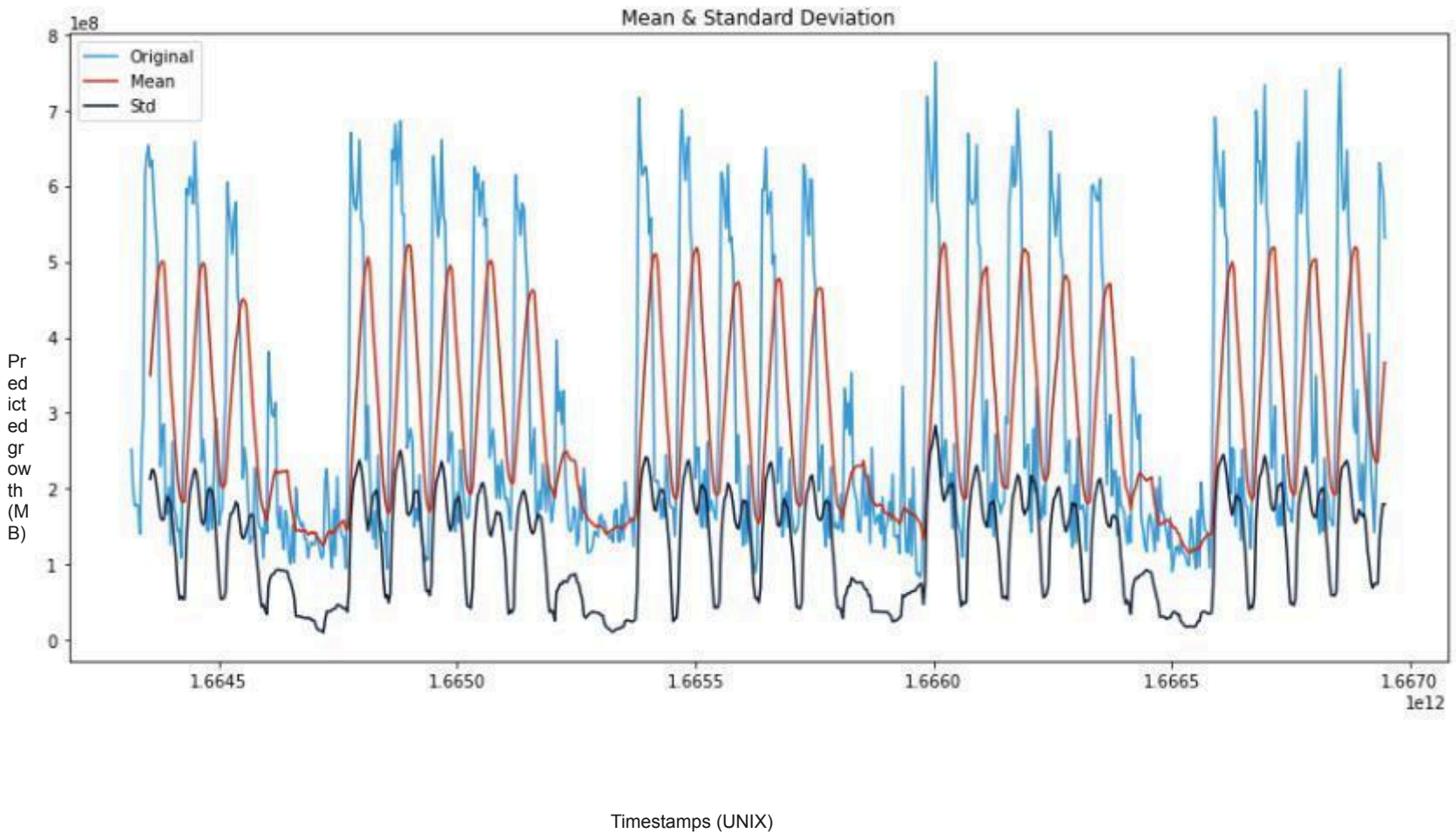

**Figure 22.** The prediction of the mean and standard deviation.

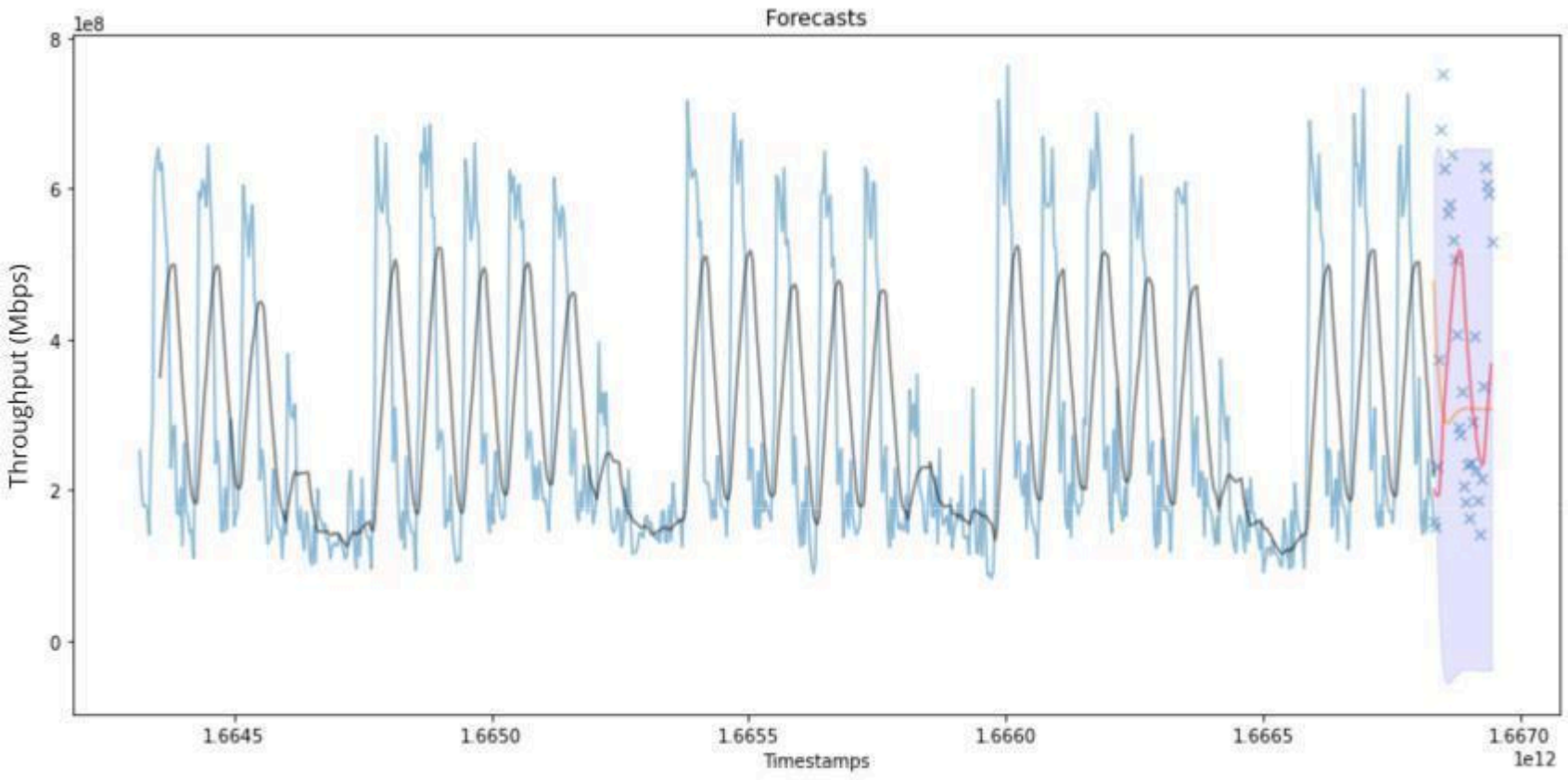

**Figure 23.** The throughput prediction.

The mean of the testing set is in red while the training set mean is black. The training set has more data points than the testing sample due to the cross-validation process. The prediction set has only 28 data points, and the forecasting is for 24 h ahead. On average, 3 Mbps is predicted for the next 24 h (Figure 23). In this study, the ARIMA code to generate the results is as follows.

**def forecast**(ARIMA model, periods = 0):

n periods = periods

**fitted**, **interval** = ARIMA model.predict(n periods = n periods, interval = True)

index = pd.date range(df.index[1], periods = n periods, freq = H)

**fitted series** = pd.Series(fitted, index = index)

**lower series** = pd.Series(interval[:, 0], index)

**upper series** = pd.Series(interval[:, 1], index)

plt.plot(fitted series)

plt.fill between(lower series.index,

lower series,

upper series,

alpha = 0.15)

plt.show()

**forecast(ARIMA model, 730)**

**ARIMA model.summary( )**

4.2. The Comparative Results of ARIMA and CNN

CNN is compared to ARIMA using the subscriber and UGRansome datasets. The CNN depends on the predicted timestamps. In what follows, we present the CNN results compared to the results obtained by the ARIMA. The prediction includes 30 to 60 days. The comparative results of the ARIMA and CNN models are in Table 7.

**Table 7.** The CNN and ARIMA results.

| Dataset | Features | p-Value | CNN | ARIMA |
|---|---|---|---|---|
| Subscriber data | 450 | 0.006, 055 | 85.8% | **92.67%** |
| UGRansome data | 120,000 | 0.0, 006, 043 | 88.9% | **91.65%** |
| Subscriber testing set | 300 | 0.008 | 86.3% | **94.8%** |
| UGRansome testing set | 60,500 | 0.007 | 88% | **95.3%** |
| **Balance** | 45,312 | 0.005 | 87.25% | **93.605%** |

The p-value is a number between 0 and 1 and can be interpreted as follows:

- A small p-value (typically <0.05) indicates strong evidence against the null hypothesis.
- A large p-value (>0.05) indicates weak evidence against the null hypothesis.

- The p-values very close to the cutoff (0.05) are considered to be marginal.

The reported p-values reject the null hypothesis stating a decrease in subscriber data usage. Moreover, CNN is compared to ARIMA using experimental datasets. We used four samples: the first sample was the subscriber dataset, where the ARIMA model obtained 92% of accuracy and outperformed the CNN. The second sample was the UGRansome dataset containing more features, but the ARIMA model surpassed the CNN with 91% of accuracy. The third sample was the testing sample of the subscriber data where the ARIMA achieved 94%. In the last sample, the ARIMA accuracy outperformed the CNN with 95% of accuracy. Overall, the ARIMA model achieved the best results in all undertaken comparisons. The ARIMA model performed better with the UGRansome data, and this was due to the nature of seasonal network traffic. We computed our models on fewer features of the subscriber data without producing poor results. We believe this is due to time series data properties which improve the balanced accuracy with 93% of accuracy.

## 4.3. Execution Speed Test

The results showed a CNN not outperforming the ARIMA in terms of accuracy, whilst the ARIMA performed better than the CNN model in terms of execution speed by a factor of 43 for more than 80,000 rows. Table 8 summarizes the speed test results for both models.

**Table 8.** Computational speed test results.

| | ARIMA (s) | CNN (s) |
|---|---|---|
| 0 rows | 0 | 0 |
| 10 rows | 0.44 | 4.39 |
| 100 rows | 0.64 | 10.43 |
| 1000 rows | 2.64 | 75.60 |
| 10,000 rows | 18.24 | 685.76 |
| 100,000 rows | 159.87 | 6951.91 |

A time Python package is used before and after each model's execution to measure how long it took to complete the entire procedure. The results are presented right after each execution is complete. The data point for each model is then plotted in batches of ten multiples (10 rows, 100 rows, 1000 rows, etc.). This gives a general notion of how each model's execution time grows as the number of data increases. Furthermore, this test is conducted using parameters for both the SARIMA and CNN models respectively. From the experimental results, it is evident that on average the ARIMA model outperforms the CNN in terms of accuracy. Table 8 showcases the speed test results for both models up to 100,000 rows. ARIMA outperforms CNN in terms of execution speed. In Table 8, which shows a linear growth for both models, the average rate of change can be determined by computing the following slope for each graph (Figure 24):

$$\therefore slope = \frac{\Delta y}{\Delta x}, \boxed{\text{ARIMA}}\, Slope = \frac{159.87 - 0}{100000 - 0} = 0.0016, \boxed{\text{CNN}}\, Slope = \frac{6951.91 - 0}{100000 - 0} = 0.069$$

**Figure 24.** The slope computation.

What can be deduced from the slopes is that on average, it takes 0.0016 s for the ARIMA model to execute one row, and 0.069 s for the CNN to execute the same row, thus making the ARIMA faster than the CNN model. Nevertheless, different parameters will yield different results. It can thus be argued that further tuning about the CNN would yield even better results.

To put it into perspective, it took under three minutes for the ARIMA to successfully execute 100,000 rows of data as compared to the CNN which took nearly two hours to complete a similar task. This margin will only widen as more rows are introduced for training between the two models. In addition, both models must be fed a sufficient amount of data to successfully train and produce the best results.

Furthermore, additional yearly data would provide findings that would boost the study's credibility in addition to pointing out parameter tuning issues for both models. Finding the ideal parameters was challenging in light of the above, especially with the CNN model. One option which was used in this research was to draw ideas from other models used in related contexts in addition to using the rule to fine-tune the model. Fortunately, this was not the case with the ARIMA model since the tuning process was relatively simple and was not based on trial and error as it was with the CNN. However, this process can further be improved for the ARIMA by a simple automated step-wise search using an Akaike Information Criterion (AIC). Developed by the Japanese statistician Hirotugu Akaike [51], AIC is used to evaluate various potential model parameters and choose the one that best fits the data. Fortunately, an AUTO ARIMA (.) function can be imported from Python's PMDARIMA library which can be used to compute the AIC for the ARIMA model. The main goal would be to develop ARIMA parameters with the lowest

AIC. For example, the ARIMA parameters for patterns in the experimental datasets were (2, 1, 2). However, if we were to perform an automated step-wise search using the AUTO ARIMA in Python (Figure 25), then the parameters would be (2, 0, 0) since it has the lowest AIC value at 103.947.

```
Performing stepwise search to minimize aic
 ARIMA(2,0,2)(0,0,0)[0] intercept   : AIC=-101.099, Time=0.11 sec
 ARIMA(0,0,0)(0,0,0)[0] intercept   : AIC=-99.831, Time=0.02 sec
 ARIMA(1,0,0)(0,0,0)[0] intercept   : AIC=-103.468, Time=0.03 sec
 ARIMA(0,0,1)(0,0,0)[0] intercept   : AIC=-101.532, Time=0.03 sec
 ARIMA(0,0,0)(0,0,0)[0]             : AIC=-55.837, Time=0.01 sec
 ARIMA(2,0,0)(0,0,0)[0] intercept   : AIC=-103.947, Time=0.05 sec
 ARIMA(3,0,0)(0,0,0)[0] intercept   : AIC=-101.948, Time=0.13 sec
 ARIMA(2,0,1)(0,0,0)[0] intercept   : AIC=-101.947, Time=0.07 sec
 ARIMA(1,0,1)(0,0,0)[0] intercept   : AIC=-103.001, Time=0.05 sec
 ARIMA(3,0,1)(0,0,0)[0] intercept   : AIC=-102.836, Time=0.15 sec
 ARIMA(2,0,0)(0,0,0)[0]             : AIC=-100.677, Time=0.03 sec

Best model:  ARIMA(2,0,0)(0,0,0)[0] intercept
Total fit time: 0.684 seconds
```

**Figure 25.** Stepwise AIC results using subscriber data.

It is thus safe to conclude that future ARIMA models can be tuned using the automated AIC stepwise search [51]. Due to the limited experience with constructing CNN models from scratch, the TensorFlow library was used to fully realize the model into practice. Unfortunately, due to the high level of abstraction of the library, it can be very challenging to realize full control over models built using TensorFlow. For example, one is unable to tune the weights of the model, thus resulting in mild changes over the final results every time the model is run due to TensorFlow's frequent changes concerning weights in the background, thus it is unclear precisely what type of network is being constructed in the background, which can result in uncertainty with regards to the performance of the model. Unfortunately, other libraries such as PyTorch and Theano exhibit similarly high levels of abstraction.

### 4.4. ARIMA, CNN, BATS, and TBATS Comparison

We have compared the obtained results with standard forecasting methods such as BATS and TBATS. BATS is an exponential smoothing technique that handles non-linear data. The advantage of using BATS is that it can treat non-linear patterns, resolve the auto-correlation issue, and account for multiple seasonality. However, BATS is computationally expensive with a large seasonal period. Hence, it is not suitable for hourly data. Thus, the TBATS model was developed to address this limitation. It represents each seasonal period as a trigonometric representation based on the Fourier series. This allows the model to fit large seasonal periods. It is thus a better choice when dealing with high-frequency data, and it usually fits faster than BATS. Figure 26 shows the comparative results with the testing sample and original subscriber data (baseline). The BATS and TBATS prediction are also illustrated in Figure 27. The TBATS outperformed the BATS with a Mean Absolute Percentage Error (MAPE) of 32.63 (Figure 27). This metric defines the accuracy of the forecasting method. The MAPE represents the average of the absolute percentage errors of each entry in a dataset to calculate how accurate the forecasted quantities were in comparison with the actual quantities. A maximum value of 6.5 KBS was predicted by the TBATS in terms of the network traffic volume (Figure 26). As such, the ARIMA model could still predict more subscriber data usage growth compared to the BATS and TBATS models. This is because ARIMA models have a fixed structure and are specifically built for time series or sequential features. It is also due to the non-modifications of predefined parameters before their implementation on time series data.

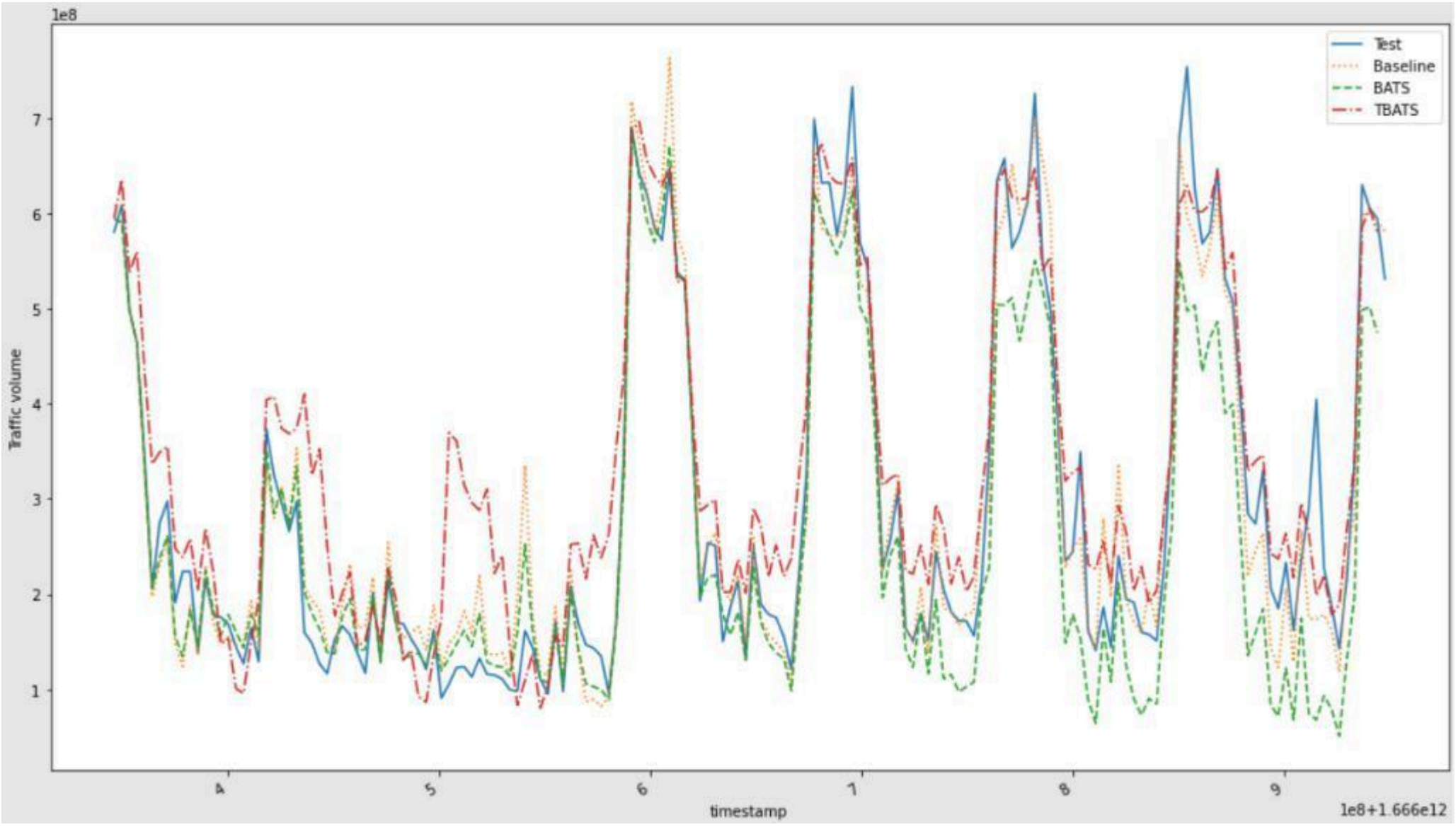

**Figure 26.** Standard forecasting approach comparison.

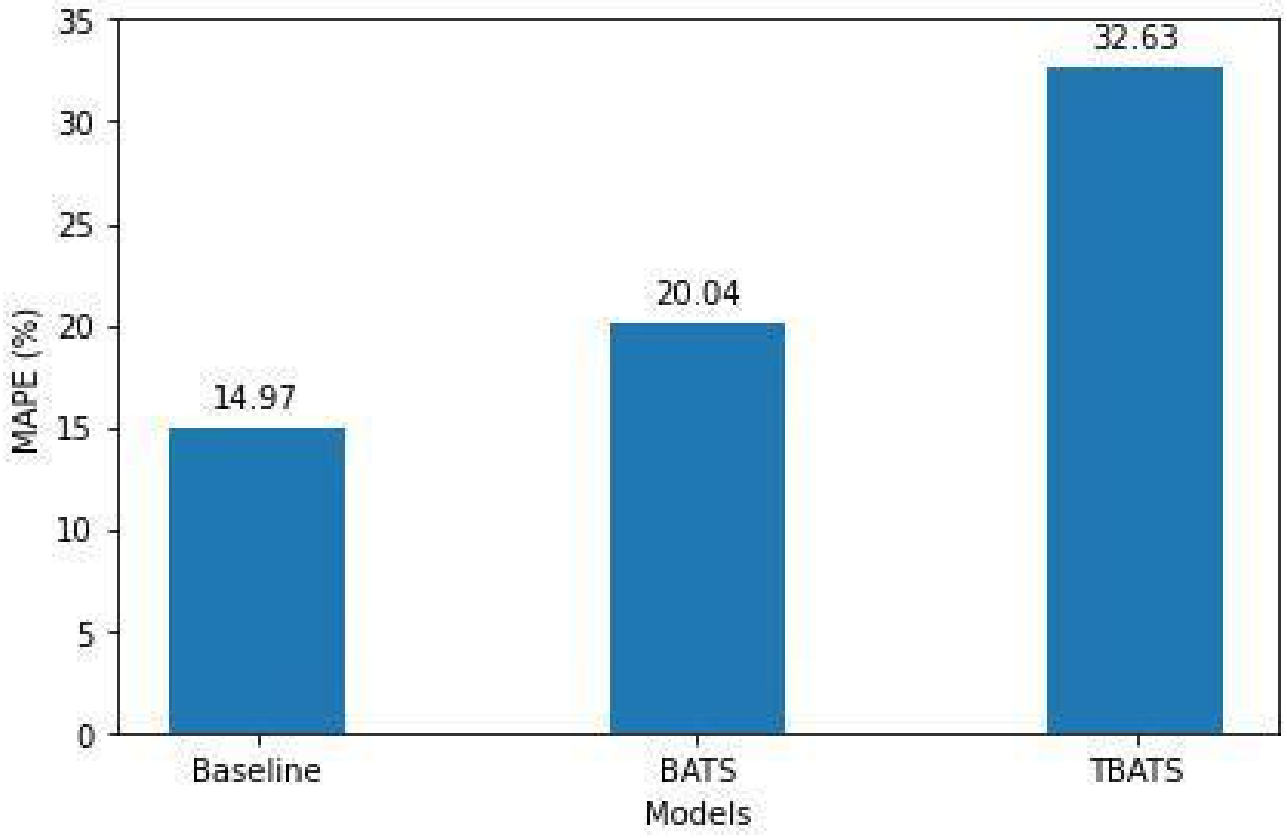

**Figure 27.** BATS and TBATS models comparison.

4.5. Recommendation

In general, the ARIMA model achieved the best predictive accuracy results on the subscriber data. With regards to model execution speed, it was posited that the ARIMA would perform better than the CNN due to the CNN's sequential weight computation for each hidden layer. However, it was not expected to outperform CNN by such a huge margin, where there was no need to validate future performance using the student t-test. Nonetheless, there are various ways to improve the execution time of CNN:

1. Python is unfortunately not the fastest language, and it is thus recommended to build ML models using low-level programming languages such as C, and C++ for faster processing times.
2. Running ML models locally via a Central Processing Unit (CPU) or Graphics Process-ing Unit (GPU) can slow down the learning process due to memory limits, and it is thus recommended to run models using cloud technologies such as Google

Colab or Amazon Web Services (AWS) and SageMaker where memory can be defined before the model is run.

3. Decreasing the number of neurons that makes up the model can reduce the processing time. However, this will be at the expense of model accuracy since fewer neurons result in model underfitting with poor performance when new data is introduced.

4. Similar to decreasing the number of neurons, decreasing the number of epochs will also reduce the final model run time, however, at the expense of accuracy since reducing the number of epochs results in underfitting the model.

## 5. Conclusions and Discussion

Insights retrieval from subscriber data impacts the telecommunication landscape to facilitate information management and assist decision-makers in predicting the future using ML techniques. We explore time series forecasting analysis and predict subscriber usage trends on the network using the ARIMA model. The unknown forecasting value used by ARIMA relied on historical data. However, we used the data storage to build the subscriber dataset using hourly traffic statistics. We used various metrics to evaluate the ARIMA model. For instance, the normal Q-Q, standardized residual, theoretical quantile, correlogram, and accuracy. UGRansome was used to compare the obtained results that demonstrate similar accuracy values of 90% using the CNN model. The subscriber data was stationary but exhibited less seasonality. In the experimentation, ARIMA was compared to CNN and achieved the best results with the UGRansome data. We have used an NSDM environment with subscriber data in a secure environment to retrieve relevant patterns such as timestamps and incoming/outgoing throughout to build the subscriber dataset. The variation of the auto-regressive and moving average components identified the most optimal features for obtaining precise predictive values. In addition, the subscriber data have normal distributions, but the UGRansome has more dependent variables. The ARIMA model predicted a growth of 3 Mbps with a maximum data usage growth of 14 Gbps. Furthermore, the performance concerning accuracy showed that ARIMA was superior to CNN. Concerning execution speed, the ARIMA outperforms the CNN by a 43:1 ratio for 100,000 rows. However, we recommend utilizing both methods depending on whether speed or accuracy is a priority. In the future, we will explore additional forecasting models by combining classical mathematical algorithms and comparing the performance with Neural Networks, specifically Recurrent Neural Networks (RNN) using ensemble learning approaches. Lastly, it will be also better to explore other factors that affect subscriber's data usage such as multivariate forecasting.

**Institutional Review Board Statement:** Not applicable.

**Informed Consent Statement:** Not applicable.

**Conflicts of Interest:** The author declares no conflict of interests.

## Abbreviations

The following abbreviations are used in this manuscript: